\documentclass[times,twocolumn,final]{elsarticle}
\usepackage{wrapfig}
\usepackage[margin=0.7in]{geometry}
\usepackage{framed,multirow}
\usepackage{booktabs}
\usepackage{tabularray}
\UseTblrLibrary{booktabs}
\usepackage{boldline} 
\usepackage{fontawesome5}
\usepackage{amssymb,amsmath}
\usepackage{romannum}
\AtBeginDocument{\pagenumbering{arabic}}
\usepackage{pifont}

\usepackage{latexsym}
\usepackage{mathrsfs}
\usepackage{float}
\usepackage{soul}
\usepackage{url}
\usepackage[dvipsnames,table,xcdraw, svgnames,x11names]{xcolor}
\usepackage{makecell}
\usepackage{subcaption}
\usepackage{algorithm}
\usepackage{algorithmic}
\usepackage{adjustbox}
\usepackage{blindtext}
\usepackage{graphicx}
\usepackage{lmodern}
\usepackage[T1]{fontenc}
\usepackage[listings,skins,breakable]{tcolorbox}
\usepackage[colorlinks]{hyperref}
\usepackage[printonlyused,withpage]{acronym}
\usepackage[nameinlink]{cleveref}
\usepackage{caption}
\usepackage{subcaption}
\usepackage[ruled, linesnumbered, vlined, algo2e]{algorithm2e}
\usepackage{algorithm}
\usepackage{algorithmic}
\usepackage{xcolor}       
\definecolor{aliceblue}{RGB}{240,248,255}

\newcommand{\appendixhead}%
{\begin{center}\textbf{{\large Appendix}}\end{center}
}
\newcolumntype{R}[2]{%
    >{\adjustbox{angle=#1,lap=\width-(#2)}\bgroup}%
    l%
    <{\egroup}%
}
\definecolor{newcolor}{rgb}{.8,.349,.1}
\definecolor{maroon}{cmyk}{0,0.87,0.68,0.32}
\definecolor{lightorange}{rgb}{1,0.753,0.478}
\newcommand{\ours}{Med-VCD}

\graphicspath{{./images/}}
\begin{document}

\begin{frontmatter}

\title{Med-VCD: Mitigating Hallucination for Medical Large Vision Language Models through Visual Contrastive Decoding}%

\author[1]{Zahra Mahdavi}
\author[2]{Zahra Khodakaramimaghsoud}
\author[3]{Hooman Khaloo}
\author[4]{Sina Bakhshandeh Taleshani}
\author[5]{Erfan Hashemi}
\author[6]{Javad Mirzapour Kaleybar}
\author[7]{Omid Nejati Manzari\corref{cor1}}

\cortext[cor1]{Corresponding author: omid\_nejaty@elec.iust.ac.ir}

\address[1]{Department of computer science, University of Central Florida, Orlando, USA}
\address[2]{Department of Bioengineering, University of Pennsylvania, Philadelphia, PA, USA}
\address[3]{Department of electrical engineering, Columbia university, New York, NY, USA}
\address[4]{Technical University of Applied Sciences Regensburg, Regensburg, Germany}
\address[5]{Department of Surgery, University of Calgary, Calgary, Alberta, Canada}
\address[6]{University College of Nabi Akram, Tabriz, Iran}
\address[7]{School of Electrical Engineering, Iran University of Science and Technology, Tehran, Iran}


\begin{abstract}
Large vision–language models (LVLMs) are now central to healthcare applications such as medical visual question answering and imaging report generation.
Yet, these models remain vulnerable to hallucination outputs that appear plausible but are in fact incorrect.
In the natural image domain, several decoding strategies have been proposed to mitigate hallucinations by reinforcing visual evidence,
but most rely on secondary decoding or rollback procedures that substantially slow inference. Moreover, existing solutions are often domain-specific and may introduce misalignment between modalities or between generated and ground-truth content.
We introduce \ours, a sparse visual-contrastive decoding method that mitigates hallucinations in medical LVLMs without the time overhead of secondary decoding.
\ours\ incorporates a novel token-sparsification strategy that selects visually informed tokens on the fly, trimming redundancy while retaining critical visual
context and thus balancing efficiency with reliability.
Evaluations on eight medical datasets, spanning ophthalmology, radiology, and pathology tasks in visual question answering, report generation,
and dedicated hallucination benchmarks, show that \ours\ raises factual accuracy by an average of 13\% and improves hallucination accuracy by 6\% relative to baseline medical LVLMs.

\end{abstract}

\end{frontmatter}

\section{Introduction}
Large Vision-Language Models (LVLMs) have achieved
significant success over the past few years \cite{Wenliang2024InstructBLIP,liu2024improved,liu2024visual}.
They have facilitated various vision and language
tasks \cite{song2024pneumollm, sun2024nodule, liu2024universal} by adapting to different input instructions. However, LVLMs are facing a grand challenge:
they often fail to accurately capture the visual content and
tend to generate fabricated responses (e.g., imaginary objects, incorrect attributes and in existent relationship), which
is known as hallucination \cite{gunjal2024detecting,liu2023mitigating}.
In medical applications, such hallucinations pose serious challenges to trustworthiness and safety, as even minor factual errors in generated reports can have critical clinical implications \cite{fan2020inf,xie2018fusing}. Consequently, enhancing factual accuracy while maintaining efficiency has become a central goal for medical LVLM deployment.

Existing hallucination mitigation methods generally fall into two categories. The first focuses on improving instruction-tuning data and re-training LVLMs to suppress hallucinations \cite{gunjal2024detecting,liu2023mitigating,lu2024evaluation,wang2024vigc}. The second category comprises Visual
Contrastive Decoding (VCD) approaches, which aim to constrain generation by contrasting logits from the original
and modified visual inputs \cite{leng2024mitigating,favero2024multi,woo2024don,manevich2024mitigating}.
While fine-tuning can enhance performance, it is often impractical in the medical domain due to limited high-quality labeled data and the domain gap between training and real-world deployment data \cite{yang2024limits}. VCD-based methods, on the other hand, typically rely on multiple decoding rounds and external visual localization tools \cite{suo2025octopus}, which introduce significant computational overhead and limit real-time usability \cite{huang2024opera, leng2024mitigating}. These limitations highlight the need for a more efficient, domain-adaptive strategy to reduce hallucinations while preserving the accuracy and reliability of medical LVLMs.

In this study, we introduce \ours\, an effective plug-and-play approach designed to mitigate hallucinations by leveraging visual-aware token sparsity in Med-LVLMs.
Motivated by empirical findings that LVLM attention patterns are inherently sparse and traditional sparsification worsens hallucinations, \ours\ employs an
innovative constrained optimization framework, a sparse-based visual contrastive decoding strategy, and Sinking Attention Calibration (SAC) to maintain visual
fidelity and efficiency. Extensive evaluations on three hallucination benchmarks and two Med-LVLM architectures confirm that \ours\ not only significantly reduces hallucinations
but also achieves superior decoding speed, demonstrating marked improvements over existing state-of-the-art methods.

The study makes several significant contributions:

\begin{itemize}
    \item We investigate hallucination reduction from the standpoint of token sparsification during decoding and introduce a new, effective, plug-and-play method
     that unites visual-aware enhancement and token sparsity as an optimization issue, achieving both model efficiency and fidelity.
    \item In order to reduce hallucination, we provide a new visual-aware token selection technique and a sparse-based VCD approach that avoids multi-round decoding
     and uses embeddings to produce contrasting logits.
    \item Extensive tests and assessments show that \ours\ performs noticeably better than current hallucination mitigation techniques in terms of both performance and decoding speed in medical domain.
\end{itemize}

\begin{figure*}
    \centering
    \includegraphics[width=0.85\linewidth]{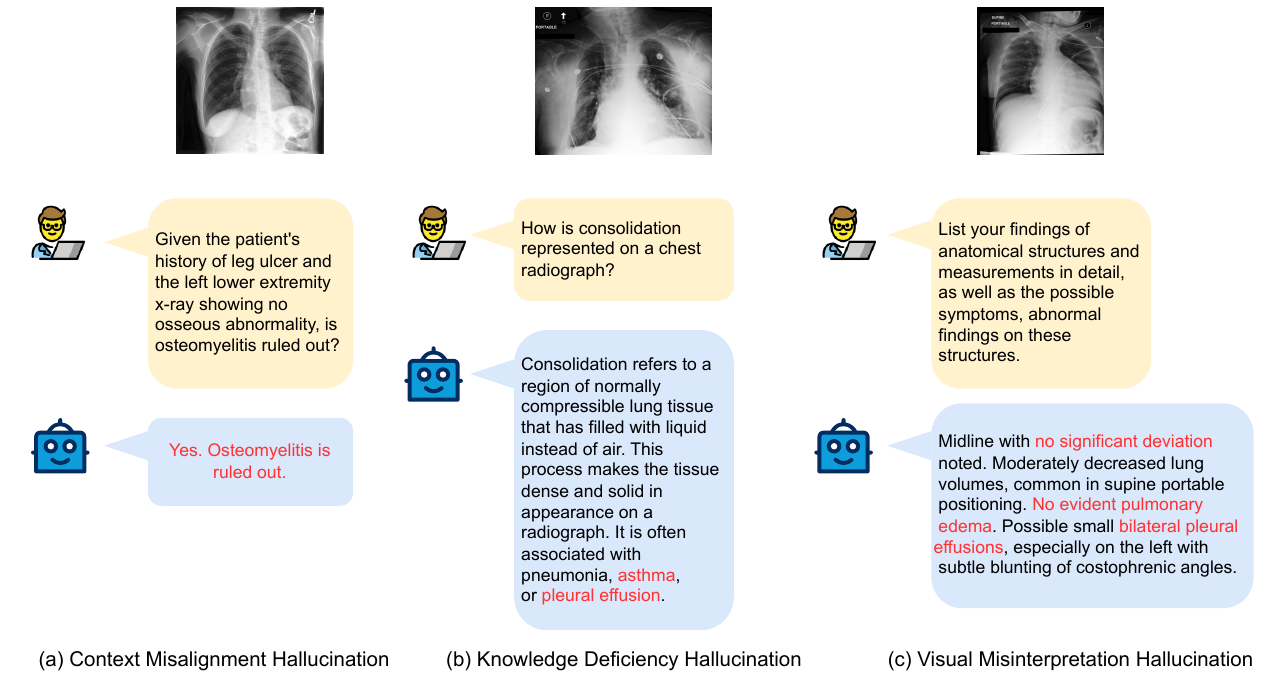}
    \caption{Illustrative cases of medical hallucination include the following:
    (a) The model incorrectly answers a context-dependent medical question; the correct response should be “No.”
    (b) The model fabricates clinical knowledge, proposing “pleural effusion” and “asthma,” whereas the appropriate diagnoses are “lung cancer” or “pulmonary edema.”
    (c) The model hallucinates the nonexistent symptom “pleural effusions” and overlooks diffuse indistinctness of the pulmonary vasculature—a radiographic finding characteristic of “pulmonary edema”.}
    \label{fig:hallucination_examples}
\end{figure*}

\section{Related Works}
\label{sec:2}

\subsection{Large-Scale Vision-Language Models in Medicine}

The development of vision-language models (VLMs) has garnered considerable attention, driven by the demonstrated success of large language models (LLMs)\cite{liu2025survey},
including GPT-4\cite{achiam2023gpt4}, LLaMA-2~\cite{touvron2023llama}, and PaLM-2~\cite{anil2023palm}. This growing momentum has catalyzed extensive research into general-domain
applications~\cite{cai2025imperceptible,liu2024visual,liu2024pandora}, resulting in frameworks such as GPT-4V~\cite{achiam2023gpt4}, PaLM-E~\cite{driess2023palm},
MiniGPT-4~\cite{zhu2023minigpt}, LLaVA~\cite{liu2024visual}, and InternVL~\cite{chen2023internvl}.

However, the development of large-scale vision–language models, particularly general-purpose foundation models, for medical applications remains relatively underexplored. Current research in medical multimodal learning generally follows two primary directions. The first direction focuses on integrating specialized medical vision models, such as Visual Med-Alpaca~\cite{shu2023visual}, OphGLM~\cite{gao2023ophglm}, and ChatCAD~\cite{wang2023chatcad}, with existing large language models~\cite{shu2023visual,gao2023ophglm,wang2023chatcad}.
These frameworks typically combine domain-specific visual encoders with LLMs,
extending capabilities to medical tasks beyond purely linguistic processing. The second direction, as highlighted by recent studies~\cite{tu2024towards,moor2023medflamingo,li2024llavamed,wu2023radfm},
emphasizes end-to-end integration, enabling joint training of vision and language modules to enhance performance in complex clinical scenarios. Examples include models such as Med-PaLM~\cite{tu2024towards},
Med-Flamingo~\cite{moor2023medflamingo}, and LLaVA-Med~\cite{li2024llavamed}, which aim to build versatile medical VLMs, alongside specialized models like RadFM~\cite{wu2023radfm},
focused on diagnosing 2D and 3D radiological imagery. In contrast to existing generalist foundation models, our proposed \ours\ framework addresses these limitations by accommodating
diverse medical imaging modalities while maintaining robust performance.

\subsection{Hallucination in Med-LVLMs}
Several benchmarks have been introduced to evaluate hallucinations in medical vision–language models (Med-LVLMs). MedVH \cite{gu2024medvh} measures robustness in visual question answering (VQA)
and report generation, Med-HallMark \cite{chen2024detecting} categorizes hallucinations by severity, and CARES \cite{xia2024cares} focuses on factual accuracy in medical VQA. More recently,
MedHallBench \cite{zuo2024medhallbench} provides automatically annotated data for assessing hallucination components. Despite recent progress, current studies are restricted by narrow datasets
and focus primarily on broad mitigation strategies, leaving limited opportunity to investigate innovative, domain-specific methods for reducing hallucinations in medical applications.

\subsection{Hallucination Mitigation}
The hallucination problem was first identified in the context of LLMs. It refers to the misalignment between LLM-generated content and real-world facts (i.e., factuality hallucination) or user
instructions (i.e., faithfulness hallucination). Building on LLMs, large vision-language models similarly suffer from hallucinations, which manifest as misalignments between generated text and
visual input \cite{kim2025medical, jiang2024hallucination, yu2024hallucidoctor}.
Despite increasing attention to hallucinations in VLMs, there remains a significant gap in research on this issue within the medical domain \cite{gu2024medvh, pham2024towards}. Existing medical
VLMs largely rely on benchmarks that do not fully capture the complexities of medical hallucinations. This limitation hinders effective mitigation and evaluation, thereby impeding the development
of reliable medical AI systems. Consequently, there is an urgent need for robust methods to enhance model performance in healthcare.

Several strategies have been proposed to address hallucinations in natural image domain, including robust instruction tuning with curated datasets \cite{gunjal2024detecting, liu2024mitigating, wu2024logical},
post-hoc methods that employ auxiliary analysis networks \cite{zhou2023analyzing, chen2024halc, wu2024logical}, and a variety of decoding approaches \cite{chuang2023dola, liu2024paying, leng2024mitigating}.
Robust instruction tuning, however, requires large-scale, high-quality datasets and substantial GPU resources, making it costly. Post-hoc methods also incur high inference costs due to their reliance on
auxiliary networks. In terms of decoding strategies, prominent LVLM hallucination mitigation methods \cite{leng2024mitigating, favero2024multi, wang2024mitigating} typically involve artificially disturbing
raw inputs to induce hallucinations and then using contrastive decoding to address them. Yet, extensive input disturbance may introduce additional noise during contrastive decoding and doubles the inference cost.

Our work focuses on designing an efficient solution for the medical domain that does not require additional training, aiming to alleviate hallucinations without the high computational overhead seen in existing methods.

\begin{figure*}
  \centering
        \includegraphics[width=0.9\textwidth]{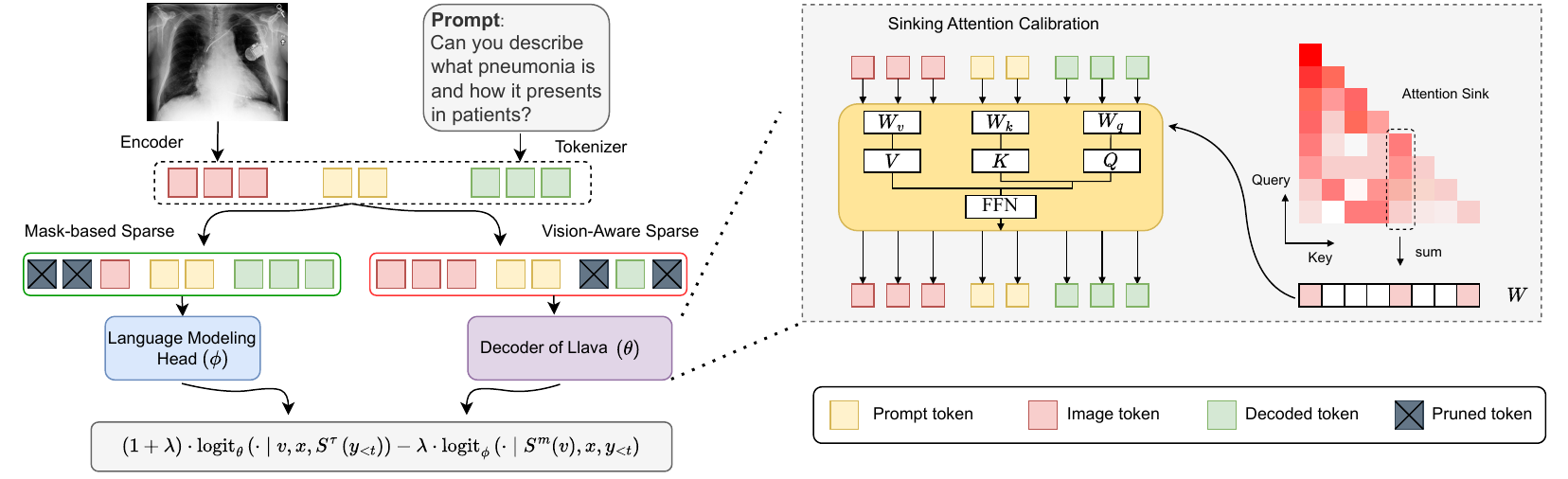}
        \captionof{figure}{An overview of the proposed \ours\ approach, consisting of (1) a sparse-based VCD method; (2) the visual-aware token selection(3); Sinking Attention Calibration.}
    \label{fig:arch}
\end{figure*}

\section{Methodology}
\label{sec:methods}

\subsection{Preliminaries}
This study investigates a general-purpose Med-LVLM, denoted as \(\theta\), which integrates a vision encoder, a LLM decoder, and a vision–text alignment module.
The vision encoder processes an input image \(v\) to generate visual embeddings, which are subsequently transformed by the alignment module, implemented using components such as a linear projection layer or a Q-Former~\cite{li2023blip}, to ensure compatibility with the textual query \(x\). The resulting multimodal representations are then passed to the decoder, which autoregressively generates the output sequence \(y\) by iteratively predicting tokens, as formalized below:

\begin{equation}
\begin{aligned}
    y_n &\sim p_{\theta}(y_n | x, v, y_{<n})\propto \exp\left( \operatorname{logit}_{\theta}(y_n | x, v, y_{<n}) \right),
\end{aligned}
\end{equation}

\noindent
here \(y_n\) denote the \(n\)-th token in the sequence \(y\), where \(y_{<n}\) represents the subsequence of tokens generated prior to the \(n\)-th step. The logit distribution
function, \(\operatorname{logit}_{\theta}\), then computes the model's output probabilities at each autoregressive generation step.

To mitigate computational redundancy, the  value \(V\) and the key \(K\) matrices within the attention mechanism are stored in a key-value cache during autoregressive decoding, as
they originate from previous decoding steps. Consequently, when generating the \(n\)-th token \(y_n\), the \(D\)-dimensional attention operation is carried out as follows:

\begin{equation}
\operatorname{Attention}(q_n, K_{\leq n}) = \operatorname{Softmax}\left(\frac{q_n K_{\leq n}^\top}{\sqrt{D}}\right),
\end{equation}

\noindent
here \(K_{\leq n}\) denotes the aggregate of keys up to (and including) the \(n\)-th decoding step, while \(q_n\) denotes the query vector for the current decoding step.

The primary goal of this study is to reduce hallucinatory token generation, thereby preserving the factual coherence of the output text while optimizing computational efficiency during autoregressive decoding.

\begin{figure}
  \centering
        \includegraphics[width=0.9\columnwidth]{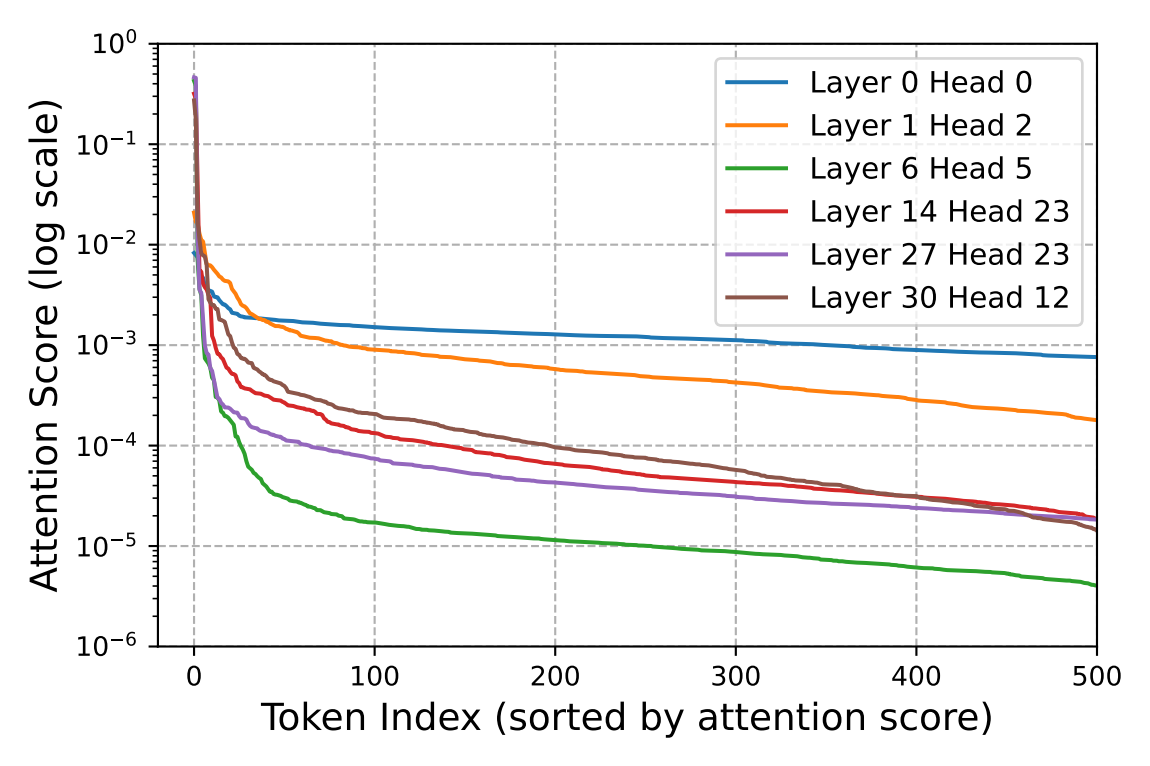}
        \captionof{figure}{Analysis token sorting by attention score using LLaVA-Med-1.5.}
    \label{fig:VCSparse}
\end{figure}

\subsection{Problem Formulation}
\label{subsec:Problem}
Attention scores associated with Med-LVLM token decoding, when sorted in ascending order (Figure \ref{fig:VCSparse}), reveal that only a small fraction of tokens receives substantial focus, yielding a
pronounced long-tail distribution. Our empirical analysis shows that most layers of the Med-LVLM decoder are governed by sparse self-attention patterns. This finding suggests a promising strategy for
improving computational efficiency during decoding by pruning tokens with negligible attention weights.
we decompose the overarching goal of improving the trustworthiness and computational efficiency of Med-LVLMs into the following methodologically grounded sub-goals:

\textbf{(1) Token Sparsification}:~
Token sparsification is formalized using a binary masking mechanism \(M\), where elements \(M_i \in \{0, 1\}\) are motivated by the inherent sparsity observed in Med-LVLMs. Let \(L\) denote the
length of sequence, with \(M_i = 0\) indicating the pruned token \(K_i\) during decoding. The goal of sparsification is to ensure that the sparse attention score \(q(K \odot M)^\top\) closely
approximates the full attention score \(qK^\top\), thus minimizing deviation. This dual objective aims to minimize the cardinality of the mask \(\sum_{i=1}^L M_i\) while preserving critical attention score distributions.

\begin{figure}
  \centering
        \includegraphics[width=0.9\columnwidth]{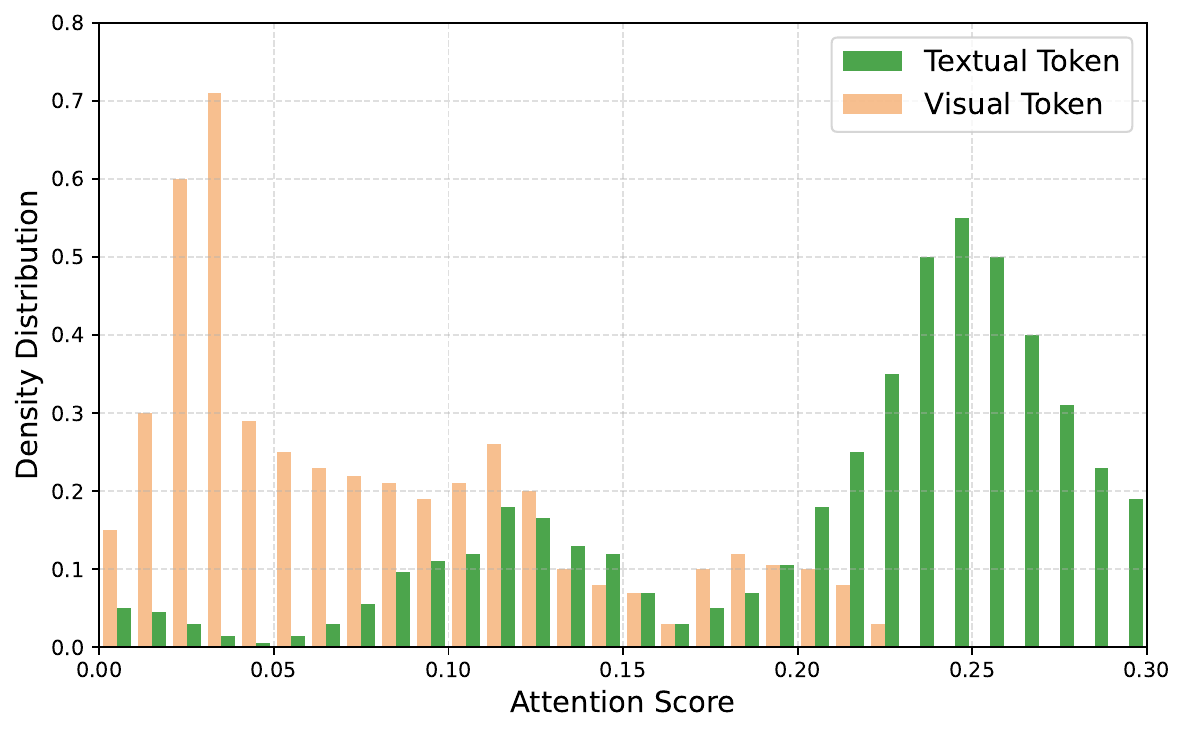}
        \captionof{figure}{Analysis attention density of textual and visual tokens using LLaVA-Med-1.5.}
    \label{fig:AttentionS}
\end{figure}

\textbf{(2) Vision-Aware Decoding}:~
Our investigations were carried out using the Open-Ended Visual Misinterpretation Benchmark and the LLaVA-Med-1.5 architecture. Evaluations were conducted on a randomly selected subset comprising 20\% of
the benchmark images. A quantitative analysis of the attention distribution patterns across image and text tokens revealed a pronounced divergence, as depicted in Figure~\ref{fig:AttentionS}. Specifically,
image tokens predominantly occupied regions of low attention, whereas text tokens showed concentrated activation within high-attention regions. Interestingly, certain tokens that received relatively low
attention weights during the decoding process nonetheless proved critical for accurately interpreting visually salient instances. These findings suggest that prematurely removing such tokens during pruning
operations may exacerbate visual hallucination phenomena, highlighting the necessity of employing cautious optimization strategies in multimodal reasoning systems. To quantify the relevance of individual
tokens when decoding visual patterns, a vision-aware saliency score \(P_i\) is assigned to each token. This score directly modulates the preservation probability during sparsification, with higher \(P_i\)
values prioritizing the retention of tokens essential for maintaining visual-semantic coherence.

The objectives outlined above can be characterized as token sparsification, systematic integration of visual-semantic cues during decoding, and minimizing deviations from the original attention score distributions.
To effectively balance these objectives, we propose a constrained optimization framework designed to simultaneously reduce the approximation error between the sparse attention scores \( q(K \odot M)^\top \)
and their full counterparts \( qK^\top \).

\subsection{Objective Definition}
\label{def:objective}
The answer to the following constrained optimization issue is what we identify as the joint aim of efficiency and trustworthiness in Med-LVLMs:
\begin{equation}\label{eq:objective}
\begin{aligned}
\min_{M} \quad \mathcal{E}(M) &= {\left\| qK^\top - q (K \odot M)^\top \right\|^2} - {M \cdot \lambda P} \\
= \sum_{i=1}^L &  \left( \langle K_i, q \rangle - {M_i} \langle K_i, q \rangle \right)^2 - M_i \cdot \lambda P_i \\
\text{s.t.} \quad M_i \in &\{0, 1\},  \forall i = 1, 2, \dots, L; \quad
\sum_{i=1}^L M_i = S,
\end{aligned}
\end{equation}
where, $K_i \in K$, $\{q, K_i\}\in \mathbb{R}^{1 \times D }$, and $||\cdot||^2$ represents the $L_2$ norm. $S$ denotes the sparsity rate, $\langle \cdot,\cdot \rangle$ is the inner product,
and the tradeoff parameter $\lambda$ is used to modify attention recall and visual perception.

The optimization framework defined in above is subject into two inherent constraints:
(1) \textbf{Visual Saliency Constraint}: Let \(P = \{P_i\}_{i=1}^L\) denote the vision-aware saliency scores, where \(P_i\) quantifies the perceptual relevance of the \(i\)-th token for visual-semantic alignment.\
(2) \textbf{Sparsity Constraint}: The binary mask \(M\) satisfies \(\sum_{i=1}^{L} M_i = S\), where \(S < L\) denotes the number of retained tokens. Each element \(M_i \in \{0,1\}\) determines
whether the \(i\)-th token is pruned.
In order to overcome these limitations, we propose a novel vision-aware token selection strategy (see Figure~\ref{fig:arch}), which enforces sparsity constraints while dynamically
prioritizing tokens crucial to visual-semantic coherence. By balancing computational frugality with fidelity to the original attention score distributions, this methodology effectively mitigates visual hallucinations.

\subsection{Visual-Aware Token Selection (VATS)}
\label{subsec:selection}
We propose a VATS strategy aimed at mitigating visual hallucinations while meeting the unified optimization objective in Equation~\ref{eq:objective}.
Specifically, for each attention head, tokens are ranked in descending order according to an aggregated saliency score \(\delta_i\). A binary mask \(M\) is then constructed by
retaining the top-\(S\) tokens \((M_i = 1)\) and pruning the rest \((M_i = 0)\). The aggregated score \(\delta_i\), which captures the joint contribution of visual saliency
and attention relevance, is formulated as follows:
\begin{equation}
    \delta_i = \left( \langle K_i, q \rangle \right)^2 + \lambda P_i,
    \label{eq:select}
\end{equation}
here, \(\langle \cdot, \cdot \rangle\) represents the inner product operator. The composite score \(\delta_i\) integrates the visual saliency metric \(P_i\) with the attention
relevance score \(\langle K, q \rangle\), ensuring the retention of visually salient tokens while preserving computational frugality. This dual integration optimizes the
trade-off between semantic fidelity to visual grounding and resource-efficient decoding.

The visual saliency scores originate from the attention scores of both produced tokens and image tokens, serving as indicators of their perceptual relevance (see Goal~2 in
Section~\ref{subsec:Problem}). Specifically, the visual saliency score \(P\) is computed using attention weights from earlier layers in the LVLM's historical computations, as formalized below:
\begin{equation}
P_i = \frac{\exp \left({\sum_{k} t_{k, i}}\right)}{\sum_{j} \exp\left({\sum_{k} t_{k, j}}\right)}; \quad
{k \in \mathcal{I}(v)},
\end{equation}
here, \(t_{i,j}\) denotes the attention score capturing the interaction between tokens \(i\) and \(j\), while \(\mathcal{I}(v)\) denotes the group of visual tokens extracted from the input image \(v\).

By leveraging precomputed attention weights between generated and image tokens, we avoid introducing additional computational overhead while preserving perceptual relevance. To process pruned tokens more efficiently, we adopt the \(k\)-nearest neighbor density peak clustering algorithm~\cite{rodriguez2014clustering} for adaptive aggregation over the pruned token set \(\mathcal{T} = \{K_i \mid M_i = 0\}\). Tokens within the same cluster are merged into a single aggregated token via weighted summation, thus compressing redundant information while retaining salient features.

\subsection{Sparse-based Visual Contrastive Decoding}
We empirically show that visual hallucinations are worsened by vision-agnostic token sparsification. However, we leverage this phenomenon constructively to mitigate language bias in
the output distribution. By comparing decoding probability distributions derived from vision-agnostic ($S^{m}$) and vision-aware ($S^{\tau}$) sparsification strategies, we propose a novel
logit redistribution approach aimed at enhancing the informative contrast within visual contexts.
Directly utilizing the output distribution from Med-LVLMs to generate a contrastive logit distribution inevitably leads to substantial computational overhead due to redundant secondary decoding
procedures. In order to overcome this challenge, we solely provide vision-agnostic token embeddings into the language decoding head $\phi$ of the language model, hence avoiding the decoder altogether.
This efficiently isolates a contrastive logit distribution at significantly lower computational cost.
More specifically, our vision-aware sparsification framework (see Section~\ref{subsec:selection}) generates the primary logit distribution ($\operatorname{logit}_{\theta}$), while stochastically
masking visual tokens and feeding their embeddings through $\phi$ yields the contrastive distribution ($\operatorname{logit}_{\phi}$). The final token probabilities are obtained through a
weighted fusion of these two distributions:

\begin{equation}
    \begin{split}
    y_n \sim (\alpha + 1) &\cdot \operatorname{logit}_\theta \left(\cdot \mid x, v, S^{\tau}(y_{<n})\right) \\
    -\alpha& \cdot\operatorname{logit}_\phi\left(\cdot \mid  x, S^{m}(v), y_{<n}\right),
    \end{split}
    \label{eq:sparse-con}
\end{equation}
here \(\alpha\) denotes a balancing parameter that governs the relative contributions of the primary and contrastive logit distributions. Crucially, our decoding framework avoids additional computational
overhead by bypassing the Med-LVLM's decoder architecture (e.g., LLaMA-V2-7B~\cite{touvron2023llama-2}), which would otherwise require a full autoregressive recomputation. Building on the methods proposed
in~\cite{Leng2023MitigatingOH}, we integrate adaptive plausibility constraints into our sparse VCD paradigm to ensure token selection remains both linguistically coherent and contextually grounded.

\begin{table*}[t]
    \centering
    \caption{Model performance (\%) of different methods based on LLaVA-Med-1.5 on medical VQA task. Notably, we report the accuracy, F1 score and AUROC. The best results and second best results are highlighted in \colorbox{red!25}{red} and \colorbox{blue!15}{blue}, respectively. }
    \resizebox{\linewidth}{!}{
    \begin{tabular}{l|cccccc|ccc|cccccc}
    \toprule
        \multirow{2}{*}{Models} & \multicolumn{6}{c|}{\textbf{Radiology}} & \multicolumn{3}{c|}{\textbf{Ophthalmology}} & \multicolumn{6}{c}{\textbf{Pathology}} \\ \cmidrule(r){2-7} \cmidrule(r){8-10} \cmidrule(r){11-16}
        & \multicolumn{3}{c}{IU-Xray} & \multicolumn{3}{c}{MIMIC-CXR} & \multicolumn{3}{|c}{Harvard-FairVLMed} & \multicolumn{3}{|c}{Quilt-1M} & \multicolumn{3}{c}{PMC-OA (Pathology)} \\ \cmidrule(r){2-4} \cmidrule(r){5-7} \cmidrule(r){8-10} \cmidrule(r){11-13} \cmidrule(r){14-16}
        & Acc & F1 & AUC & Acc & F1 & AUC & Acc & F1 & AUC & Acc &F1 &AUC & Acc &F1& AUC \\
        \midrule
        LLaVA-Med-1.5 & 75.47 & 64.04 & 67.46 & 75.79 & 80.49 & 68.84 & 63.03&74.11&63.05 & 62.80&72.90&60.03 & 59.28&71.98&54.19  \\
        \midrule
        + Greedy & 76.88 & 65.59 & 68.74 & 78.32 & 86.75 & 71.13 & 82.54&85.98&70.09 & 64.72&70.12&58.75 & 58.61&70.42&53.10 \\
        + Beam Search & 76.91 & 66.06 & 68.77 & 81.56 & 86.36 & 73.79 & 80.93&88.08&68.94 & 63.52&69.33&57.65 & 56.29&69.84&52.89 \\
        + DoLa & 78.00 & 66.75 & 72.19 & 81.35  & 85.73 & 72.73 & 76.87&85.53&67.10 & 63.47&69.10&57.58 & 57.71&70.27&52.95 \\
        + OPERA & 70.59 & 61.54 & 63.22 & 69.34 & 76.66 & 62.46 & 71.41&81.37&65.59 & 60.51&66.32&54.79 & 55.32&68.30&51.86 \\
        + VCD & 68.99 & 54.35 & 61.08 & 70.89 & 75.57 & 64.61 & 65.88&77.20&64.16 & 61.43&67.39&55.72 & 55.10&67.94&51.62 \\ \midrule
        + MedDr & 83.33 & 67.80 & 77.15 & 55.16 & 56.18 & 58.47 & 70.17&80.72&64.15 & 68.15&73.23&67.01 & 59.97&69.19&57.01 \\
        + FactMM-RAG & 84.51 & 68.51 & 77.07 & 77.58 & 81.86 & 70.09 & 83.67 & 87.21 & 72.20 & \cellcolor{blue!15}{69.25} & 73.62 & \cellcolor{blue!15}{68.15} & 60.49 & 69.38 & 57.31 \\
        + RULE & \cellcolor{blue!15}{87.84} & \cellcolor{blue!15}{78.00} & \cellcolor{blue!15}{85.78} & \cellcolor{red!25}{83.92} & \cellcolor{blue!15}{87.49} & \cellcolor{blue!15}{83.44} & \cellcolor{blue!15}{87.12} & \cellcolor{red!25}{92.89}& \cellcolor{blue!15}{77.08} & 68.97 & \cellcolor{blue!15}{73.80}& 68.13 & \cellcolor{blue!15}{61.41} & \cellcolor{blue!15}{70.36} & \cellcolor{blue!15}{58.91} \\
        \midrule
        Ours & \cellcolor{red!25}{90.56}& \cellcolor{red!25}{80.30}& \cellcolor{red!25}{88.03} & \cellcolor{blue!15}{82.95}& \cellcolor{red!25}{89.38}&
        \cellcolor{red!25}{85.42} &
        \cellcolor{red!25}{88.73} &
        \cellcolor{blue!15}{92.58} &
        \cellcolor{red!25}{81.36} & \cellcolor{red!25}{73.52} & \cellcolor{red!25}{75.60} & \cellcolor{red!25}{72.96} & \cellcolor{red!25}{65.31} & \cellcolor{red!25}{74.19} & \cellcolor{red!25}{62.38} \\
    \bottomrule
    \end{tabular}
    }
    \label{tab:vqa}
\end{table*}

\vspace{1em}

\begin{table*}[t]
    \centering
    \caption{Model performance (\%) of different methods on report generation task. Notably, we report the average BLEU score, ROUGE-L, METEOR.}
    \resizebox{\linewidth}{!}{
    \begin{tabular}{l|cccccc|ccc}
    \toprule
        \multirow{2}{*}{Models} & \multicolumn{6}{c|}{\textbf{Radiology}} & \multicolumn{3}{c}{\textbf{Ophthalmology}} \\ \cmidrule(r){2-7} \cmidrule(r){8-10}
        & \multicolumn{3}{c}{IU-Xray} & \multicolumn{3}{c}{MIMIC-CXR} & \multicolumn{3}{c}{Harvard-FairVLMed}  \\ \cmidrule(r){2-4} \cmidrule(r){5-7} \cmidrule(r){8-10}
        & BLEU & ROUGE-L & METEOR & BLEU & ROUGE-L & METEOR & BLEU & ROUGE-L & METEOR \\
        \midrule
        LLaVA-Med-1.5 & 9.64&12.26&8.21 &  12.11&13.05&11.16 & 18.11&11.36&10.75  \\
        \midrule
        + Greedy & 11.47&15.38&12.69 & 16.63&14.26&14.19 & 17.98&11.49&13.77 \\
        + Beam Search & 12.10&16.21&13.17 & 16.97&14.74&14.43 & 18.37&12.62&14.50 \\
        + DoLa & 11.79&15.82&12.72 & 17.11&14.89&14.81 & 18.26&12.51&14.51 \\
        + OPERA & 10.66&14.70&12.01 & 15.40&12.52&13.72 & 16.59&11.47&13.63  \\
        + VCD & 10.42&14.14&11.59 & 15.18&12.30&13.38 & 16.73&11.38&13.89  \\ \midrule
        + MedDr & 12.37&16.45&13.50 & 18.59&15.72&16.77 & 19.82&13.72&15.40 \\
        + FactMM-RAG & 14.70 & 18.05 & 15.92 & \cellcolor{blue!15}{18.71} & {15.84} & 16.82 & 20.82 & 14.17 & 15.31 \\
        + RULE & \cellcolor{blue!15}{27.53} & \cellcolor{blue!15}{23.16} & \cellcolor{blue!15}{27.99} & 18.61& \cellcolor{red!25}{15.96}& \cellcolor{blue!15}{17.42} & \cellcolor{blue!15}{22.35} & \cellcolor{blue!15}{14.93} & \cellcolor{blue!15}{17.74} \\
        \midrule
        Ours & \cellcolor{red!25}{32.59} & \cellcolor{red!25}{25.63}&\cellcolor{red!25}{31.20} & \cellcolor{red!25}{24.12} & \cellcolor{blue!15}{15.89} & \cellcolor{red!25}{21.95} & \cellcolor{red!25}{23.15} & \cellcolor{red!25}{16.79}&\cellcolor{red!25}{18.13} \\
    \bottomrule
    \end{tabular}
    }
    \label{tab:report}
\end{table*}

\subsection{Sinking Attention Calibration}
\label{subsec:attn}
Several studies \cite{leng2024mitigating, huang2024opera, huo2024self} highlights a pronounced attention sinking phenomenon in LVLMs, where certain tokens receive disproportionately high attention
scores despite minimal semantic relevance to the visual context. During decoding, excessive focus on these tokens can overshadow critical visual information. In order to satisfy this, we propose a targeted
penalty mechanism for tokens exhibiting anomalous attention patterns. A learnable penalty weight matrix \(W = \{w_1, w_2, \ldots, w_L\}\) is introduced, where each \(w_i \in W\) dynamically scales the
attention score of the \(i\)-th token to suppress over-activation.
To quantify the severity of attention sinking, we compute each token's cumulative attention score by aggregating attention contributions across sequential queries. These aggregated scores are
subsequently normalized via a softmax operation to yield calibrated penalty weights, expressed as:
\begin{equation}
w_j = \frac{\exp\left( \sum_{i=j}^{L} a_{i,j} \right)}{\sum_{k=1}^{L} \exp\left( \sum_{i=k}^{L} a_{i,k} \right)},
\label{eq:penilty}
\end{equation}
here, \(w_j \in W\) denotes the \(j\)-th component of the normalized weight vector following the \(\mathrm{softmax}\) function, and \(a_{i,j}\) denotes the attention score between the \(i\)-th
query and the \(j\)-th key in the attention matrix. As illustrated in Figure~\ref{fig:arch}, this approach ensures progressive evaluation of attention sinking across sequential decoding steps.
The penalty weight matrix \(W\) is incorporated into the attention computation via the transformation:
$(1 + \beta)\, qK^{\top} \;-\; \beta \,\bigl(W \odot qK^{\top}\bigr),$
where \(\beta\) modulates the penalty intensity. This formulation dynamically suppresses over-activation of semantically sparse tokens while preserving attention fidelity to visually salient patterns.

%
%

\section{Experiment}
This section evaluates the performance of \ours\ by addressing four specific objectives: Assess whether \ours\ improves the factuality of Med-LVLMs compared to decoding-based and RAG-based baselines.
Quantify the individual contribution of each proposed component to the overall performance. Evaluate the effectiveness of our proposed approach in mitigating hallucinations across three medical benchmarks.
Verify that \ours\ enhances both cross-modality alignment and overall model alignment.

\subsection{Experimental Setups}
\label{sec:exp}
\textbf{Implementation Details}.
We use LLaVA-Med and LLaVA-Med-1.5 7B~\citep{li2023llava} as the backbone model. We implement the \ours\ using the Transformer library \cite{wolf2019huggingface} and
integrate it with beam search for the decoding process. Our evaluation utilizes a maximum generation sequence length of 1024 tokens, with the beam size configured to 2.
The sparsity rate is established as 0.8 times the sequence length, while the image masking sparsity rate is set to 0.5. The
hyperparameters $\beta$, $\alpha$, and $\lambda$ are configured to 0.1, 0.3, and 0.1, respectively. The decoding procedure for the Med-LVLMs and all related experiments are conducted on A100 GPUs.

\textbf{Baseline Methods}. Two types of LVLM hallucination reduction techniques that have encouraging outcomes in natural image understanding are compared to \ours . \\
1) Decoding-based techniques, such as VCD~\citep{leng2024mitigating}, OPERA~\citep{huang2024opera}, DoLa~\citep{chuang2023dola}, Beam Search~\citep{sutskever2014sequence},
and Greedy Decoding. These methods manipulate the logits of the model's
output tokens to enhance factual accuracy.\\
2)  Multimodal RAG-based methods, including RULE~\citep{xia2024rule}, FactMM-RAG~\citep{sun2024fact}, and MedDr~\citep{he2024meddr}. Additionally, we evaluate the performance against other
open-source Med-LVLMs, such as RadFM~\citep{wu2023towards}, MedVInT~\citep{zhang2023pmc}, and Med-Flamingo~\citep{moor2023med}.

\textbf{Evaluation Datasets}.
\label{sec:dataset}
We utilize five medical vision-language datasets for medical VQA and report generation tasks, i.e., IU-Xray~\citep{demner2016preparing}, MIMIC-CXR~\citep{johnson2019mimic}, Quilt-1M~\citep{ikezogwo2023quilt},
PMC-OA~\citep{lin2023pmc} (we only select the pathology part) and Harvard-FairVLMed~\citep{luo2024fairclip}. Pathology, ophthalmology, and radiology are all covered by these databases. In order to create the VQA
benchmarks, as per ~\citep{xia2024cares}, we use GPT-4~\citep{achiam2023gpt} to create question-answer pairs from medical reports, with responses structured as \textit{yes} or \textit{no}. Because pathology photos
have short and inadequate explanations, they are not included in the report production process.

We evaluate the effectiveness of our \ours\ in hallucination mitigation of Med-LVLMs on three medical benchmarks \cite{chang2025medheval}: (1) visual misinterpretation hallucination separate the evaluation into two distinct datasets include,
Multi-Modality Visual Hallucination (MM-VisHal) and Chest X-ray Visual Hallucination (CXR-VisHal). (2) knowledge deficiency hallucination data are constructed based on the MIMIC-CXR test set, where the
imaging report is used as the imaging interpretation to prompt GPT-4 to extract and construct diagnostic questions. (3) context misalignment hallucination link MIMIC-CXR data with a de-identified EHR
dataset MIMIC-IV \cite{johnson2023mimic} by the subject ID to provide comprehensive medical notes for the chest X-rays for each individual.

\textbf{Evaluation Metrics}.
Following~\citep{xia2024rule,lin2023medical}, we use Accuracy, F1 Score and AUROC for evaluating medical VQA task, and BLEU Score~\citep{papineni2002bleu}, ROUGE-L~\citep{lin2004rouge} and
METEOR~\citep{banerjee2005meteor} for evaluating report generation task. In alignment with existing hallucination benchmarks in both general and medical domains, accuracy (Acc) is employed as
the primary metric for evaluating close-ended hallucination. Assessing open-ended hallucinations in generated reports, we follow CheXpert~\cite{irvin2019chexpert} and measure hallucination rates
using CHAIR~\cite{li-etal-2023-evaluating}, which evaluates key symptom-centered visual findings. CHAIR is defined as:
$
\text{CHAIR} = \frac{|\mathcal{G} - \mathcal{S}|}{|\mathcal{G}|},
$
where $\mathcal{G}$  represents the set of findings extracted from the generated report using CheXbert~\cite{smit2020combining}, and $\mathcal{S}$ represents the set of findings extracted from the
real report using the same method.
For a more comprehensive evaluation, we additionally report key findings recall (Recall) and assess overall report quality using specialized metrics such as CheXbert~\cite{smit2020combining},
RadGraph~\cite{jain2021radgraph}, and RaTEScore~\cite{zhao2024ratescore}, which have been specifically developed for medical report generation. These metrics align closely with radiologists' assessments,
making them particularly suitable for evaluating the generation of open-ended medical reports, as demonstrated by RaTEScore data.

\subsection{Comparison with Baselines.}
On medical VQA and report production tasks, we compare our approaches with baseline methods. The results are shown in Table~\ref{tab:vqa} and Table~\ref{tab:report}, respectively.
All things considered, \ours\ performs better than all baselines in almost every measure and dataset.  In particular, \ours\ shows a notable improvement in performance, outperforming
the original Med-LVLM by 25.7\% and 22.9\% in medical VQA and report generating tasks, respectively.  With improvements of 22.8\% and 22.2\% in the two tasks, \ours\ outperforms decoding-based
methods when compared to baseline methods.
Furthermore, our method beats RAG-based baselines by 5\% and 17.8\% in the medical VQA and report creation tasks, respectively, despite recent RAG-based approaches showing significant gains over
prior techniques. This suggests that \ours\ successfully reduces the misalignment problems brought on by VCD.  Interestingly, \ours\ generates reports with more noticeable improvements, probably
because the task is more complex and retrieved contexts have a bigger impact on open-ended generation.

\begin{table}[!t]
    \centering
    \caption{Performance comparison with several Med-LVLMs. Opt: Ophthalomology, Rad: Radiology, Pat: Pathology.}
    \label{tab:other}
    \begin{tabular}{l|ccc}
    \toprule
    Model & Rad & Opt & Pat \\\midrule
    Med-Flamingo & 27.42 & 22.50 & \underline{29.11}  \\
    MedVInT  & 33.17 & \underline{29.40} & 25.33 \\
    RadFM & 35.82 & 27.07 & 24.82 \\
    miniGPT-Med & \underline{36.66} & 25.28 & 23.16 \\
    Ours & \textbf{57.74} & \textbf{53.90} & \textbf{55.39}  \\
    \bottomrule
    \end{tabular}
\end{table}

\subsection{Comparison with Other Med-LVLMs.} To show the superiority of our method, we compare \ours\ to other open-source Med-LVLMs in order to give a thorough
comparison. We evaluate these models' performance in several medical image modalities, and Table~\ref{tab:other} shows the average outcomes for medical VQA and
report generating tasks.  Our results demonstrate that \ours\ performs noticeably better than Med-LVLMs that have been pre-trained on extensive datasets in a variety of domains.
This demonstrates our method's efficacy and generalizability across many image domains and medical multimodal tasks.

\begin{table*}[t]
\centering
\caption{Performance of hallucination mitigation methods on closed-ended visual misinterpretation hallucination evaluation, applied to LLaVA-Med and LLaVA-Med-1.5.}
\label{tab:close_ended_VFH_mitigation}
 \resizebox{0.9\textwidth}{!}
{
\begin{tabular}{l|cccc|c|cccc|c}
\toprule
\multirow{2}{*}{\textbf{Models}} & \multicolumn{5}{c|}{\textbf{MM-VisHal}} & \multicolumn{5}{c}{\textbf{CXR-VisHal}} \\
\cline{2-11}
 & \textbf{Acc-A $\uparrow$} & \textbf{Acc-M$ \uparrow$} & \textbf{Acc-S $\uparrow$} & \textbf{Acc-R $\uparrow$} & \textbf{Acc $\uparrow$}
  & \textbf{Acc-A $\uparrow$} & \textbf{Acc-M $\uparrow$} & \textbf{Acc-S $\uparrow$} & \textbf{Acc-R $\uparrow$} & \textbf{Acc $\uparrow$} \\
\midrule

LLaVA-Med & 0.525 & 0.357 & 0.584 & 0.485 & 0.499 & 0.698 & 0.452 & 0.725 & 0.800 & 0.698 \\
\midrule
+ VCD & 0.529 & 0.348 & 0.578 & 0.489 & 0.496 & 0.714 & 0.448 & 0.723 & 0.786 & 0.697 \\
+ DoLa & 0.530 & 0.361 & 0.591 & 0.495 & 0.505 & 0.731 & 0.466 & 0.754 & 0.802 & 0.723 \\
+ OPERA & 0.531 & 0.352 & 0.588 & 0.508 & 0.505 & 0.749  & 0.499 & 0.757 & 0.813 & 0.732 \\
+ AVISC & 0.534 & 0.362 & 0.577 & 0.462 & 0.496 & 0.703 & 0.444 & 0.743 & 0.835 & 0.712 \\
+ M3ID & 0.529 & 0.376 & 0.585 & 0.497 & 0.507 & 0.691 & 0.426 & 0.721 & 0.804 & 0.691 \\
+ DAMRO & 0.517 & 0.369 & 0.600 & 0.459 & 0.501 & 0.698 & 0.439 & 0.728 & 0.821 & 0.700 \\
+ PAI & 0.537 & 0.359 & 0.588 & 0.507 & 0.507 & 0.745 & 0.450 & 0.757 & 0.805 & 0.726 \\
\rowcolor{aliceblue} Ours & 0.578 & 0.401 & 0.598 & 0.513 & 0.524 & 0.771 & 0.528 & 0.769 & 0.840 & 0.757 \\
\midrule
\midrule
LLaVA-Med-1.5 & 0.619 & 0.397 & 0.499 & 0.483 & 0.499 & 0.840 & 0.494 & 0.651 & 0.845 & 0.684 \\
\midrule
+ VCD & 0.610 & 0.386 & 0.489 & 0.482 & 0.491 & 0.819 & 0.489 & 0.637 & 0.845 & 0.672 \\
+ DoLa & 0.633 & 0.387 & 0.500 & 0.500 & 0.503 & 0.866 & 0.491 & 0.688 & 0.872 & 0.714 \\
+ OPERA & 0.636 & 0.387 & 0.520 & 0.515 & 0.514 & 0.856 & 0.530 & 0.729 & 0.872 & 0.742 \\
+ AVISC & 0.642 & 0.376 & 0.492 & 0.495 & 0.499 & 0.837 & 0.489 & 0.623 & 0.837 & 0.665 \\
+ M3ID & 0.616 & 0.361 & 0.511 & 0.512 & 0.499 & 0.827 & 0.473 & 0.633 & 0.831 & 0.667 \\
+ DAMRO & 0.626 & 0.360 & 0.489 & 0.493 & 0.490 & 0.833 & 0.493 & 0.605 & 0.833 & 0.654 \\
+ PAI & 0.640 & 0.385 & 0.528 & 0.495 & 0.514 & 0.866 & 0.491 & 0.715 & 0.870 & 0.731 \\
\rowcolor{aliceblue} Ours & 0.671 & 0.403 & 0.554 & 0.528 & 0.549 & 0.883 & 0.539 & 0.755 & 0.886 & 0.766 \\
\bottomrule
\end{tabular} }
\end{table*}

\subsection{Visual Misinterpretation Hallucination}
A visual misinterpretation hallucination occurs when the model interprets fundamental visual components that are factually incorrect
or unsupported by medical evidence.

\noindent
\textbf{Hallucination Mitigation on Closed-Ended. } \\
To further evaluate the effectiveness of hallucination mitigation techniques in LVLMs, we apply several existing methods to LLaVA-Med and LLaVA-Med-1.5. The comparative results are summarized in
Table~\ref{tab:close_ended_VFH_mitigation}. Overall, techniques such as PAI, OPERA, and DoLa consistently reduce hallucinations across multiple subsets, highlighting their robust performance.
However, individual methods exhibit varying strengths depending on the evaluation subset. For instance, DAMRO performs poorly on MM-VisHal when applied to LLaVA-Med but surpasses all baseline
methods on Acc-S. This variation underscores the complexity of hallucinations, emphasizing the need for specialized and context-aware mitigation strategies. Notably, our \ours\ consistently
achieves superior results across most evaluation settings, outperforming state-of-the-art post-processing and decoding methods. Furthermore, \ours\ markedly reduces hallucinations across two
distinct Med-LVLM architectures, demonstrating both its broad generalizability and its plug-and-play versatility.

\begin{table}[t]
\centering
\caption{Results of hallucination mitigation methods on open-ended visual misinterpretation evaluation.}
\label{tab:type1_open_mitigation}
\resizebox{\columnwidth}{!}{
\begin{tabular}{l|c|c|c|c|c}
\hline
\textbf{Models}  & \textbf{CheXbert $\uparrow$} & \textbf{RadGraph $\uparrow$} & \textbf{RaTEScore $\uparrow$}& \textbf{Recall $\uparrow$} & \textbf{CHAIR $\downarrow$} \\
\hline
LLaVA-Med & 19.72 & 7.31 & 39.86 & 25.17 & 20.85 \\
\midrule
+ VCD & 19.64 & 7.16 & 39.89 & 24.23 & 20.15   \\
+ DoLa & 20.68 & 7.09 &  41.11 & 29.00 & 24.31 \\
+ OPERA & 23.59 & 8.77 & 41.41 & 27.55 & 16.39 \\
+ AVISC & 20.48 & 7.79 & 40.76 & 31.29 & 22.21 \\
+ M3ID & 19.48 & 7.55 & 40.29 & 22.87 & 20.69 \\
+ DAMRO & 20.99 & 7.87 & 40.96 & 31.21 & 19.30  \\
+ PAI & 19.94 & 7.09 & 41.20 & 29.51 & 27.31 \\
\rowcolor{aliceblue} Ours & 24.50 & 8.97 & 41.33 & 32.74 & 15.02 \\
\midrule
\midrule
LLaVA-Med-1.5 & 18.44 & 4.96 & 39.47 & 13.27 & 19.74 \\
\midrule
+ VCD & 17.89 & 5.15 & 39.80 & 13.95 & 21.06  \\
+ DoLa & 20.53 & 7.48 &  39.11 & 6.29 & 23.88 \\
+ OPERA & 18.64 & 5.01 & 41.44 & 19.13 & 20.98 \\
+ AVISC & 16.12 & 4.08 & 40.34 & 11.82 & 19.16 \\
+ M3ID & 18.77 & 4.03 & 38.93 & 10.71 & 18.18 \\
+ DAMRO & 18.08 & 4.96 & 39.86 & 14.88 & 18.28 \\
+ PAI & 23.13 & 9.09 & 37.87 & 7.82 & 24.12 \\
\rowcolor{aliceblue} Ours & 23.81 & 8.29 & 40.76 & 20.37 & 17.05 \\
\hline
\end{tabular}}
\end{table}

\begin{figure*}[htp]
\centering
    \subfloat[Hallucination mitigation on LLaVA-Med]{%
        \includegraphics[width=0.48\linewidth]{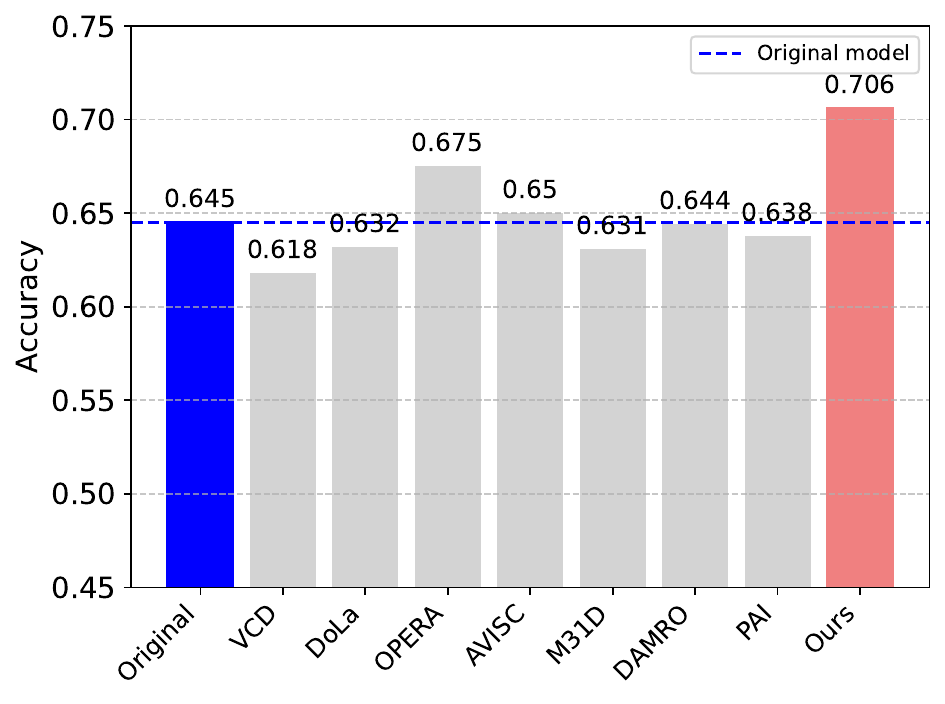}%
        \label{fig:deficiency_a}%
        }%
    \hfill%
    \subfloat[Hallucination mitigation on LLaVA-Med-1.5]{%
        \includegraphics[width=0.48\linewidth]{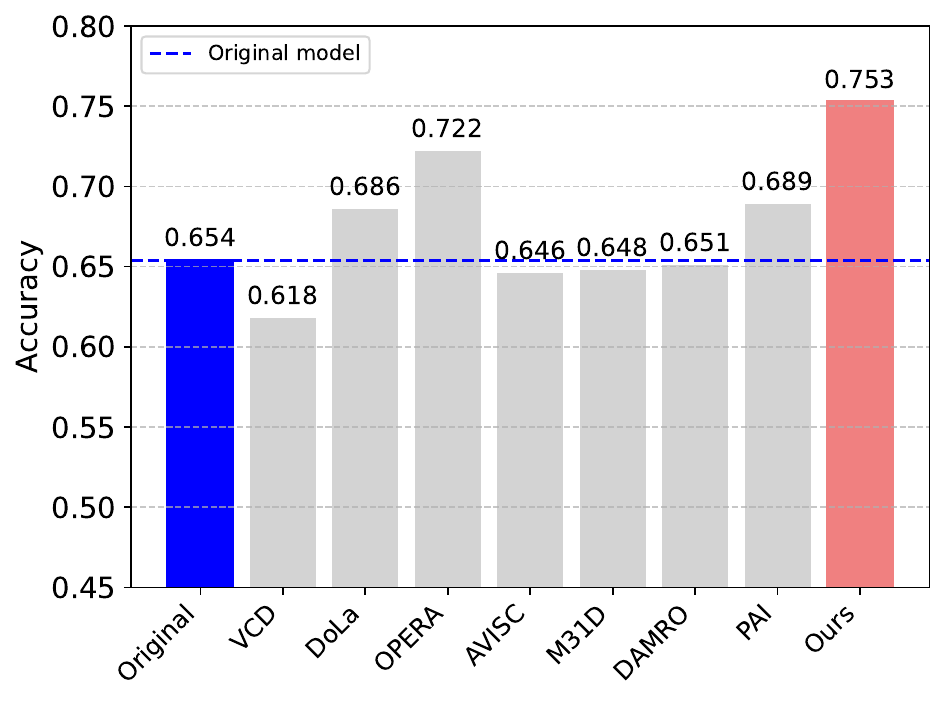}%
        \label{fig:deficiency_b}%
        }%
    \caption{Close-ended evaluation of knowledge deficiency hallucination in (Med)-LVLMs and the effectiveness of hallucination
mitigation methods.}
\end{figure*}

\begin{table*}[htp]
\centering
\caption{Results of hallucination mitigation methods on open-ended knowledge deficiency hallucination evaluation.}
\label{tab:open_ended_KH_mitigation}
\resizebox{0.9\textwidth}{!}{
\begin{tabular}{l|cccccc|c}
\toprule
\multirow{2}{*}{\textbf{Models}} & \multicolumn{6}{c|}{\textbf{Generation Metrics}} & \multicolumn{1}{c}{\textbf{Hallucination Score}} \\
\cline{2-8}
 & \textbf{BertScore $\uparrow$} & \textbf{BLEU $\uparrow$} & \textbf{METEOR $\uparrow$} & \textbf{ROUGE-1 $\uparrow$}
  & \textbf{ROUGE-2 $\uparrow$} & \textbf{ROUGE-L $\uparrow$} & \textbf{$\mathcal{S}_h$ $\downarrow$} \\
\midrule
LLaVA-Med & 89.51 & 11.59 & 40.68 & 41.44 &  19.99 & 30.44 & 1.92 ± 1.02  \\
\midrule
+ VCD & 89.54 & 11.70 & 40.63 & 41.66 &  20.09 & 30.61 & 1.96 ± 1.16 \\
+ DoLa & 89.61 & 11.93 & 41.09 & 41.90 &  20.45 & 30.93 & 1.84 ± 1.09\\
+ OPERA & 88.60 & 9.94 & 34.02 & 38.78 &  18.53 & 28.95 & 1.99 ± 1.45  \\
+ AVISC & 89.39 & 11.12 & 40.55 & 41.09 &  19.43 & 29.85 & 2.06 ± 1.13 \\
+ M3ID & 89.50 & 11.53 & 40.75 & 41.50 &  19.94 & 30.40 & 2.05 ± 1.18  \\
+ DAMRO & 89.43 & 11.30 & 40.72 & 41.18 & 19.70 & 30.06 & 2.07 ± 1.18 \\
+ PAI & 89.71 & 12.41 & 41.42 & 42.63 &  20.97 & 31.54 & 1.76 ± 1.16 \\
\rowcolor{aliceblue} Ours & 90.09 & 14.61 & 42.17 & 43.94 &  21.14 & 32.18 & 1.63 ± 1.13 \\
\midrule
\midrule
LLaVA-Med-1.5 & 89.61 & 12.14 & 40.77 & 42.93 & 20.49 & 31.29 & 1.76 ± 0.96 \\
\midrule
+ VCD & 89.60 & 11.87 & 40.46 & 42.68 & 20.19 & 30.99 & 1.79 ± 1.06 \\
+ DoLa & 89.82 & 12.85 & 41.43 & 43.68  & 21.29 & 32.14 & 1.63 ± 1.02 \\
+ OPERA & 89.62 & 13.01 & 41.45 & 43.78  & 21.48 & 32.25 & 1.58 ± 0.93 \\
+ AVISC & 89.44 & 11.14 & 40.23 & 42.22  & 19.39 & 30.25 & 1.82 ± 0.97 \\
+ M3ID & 89.50 & 11.69 & 40.09 & 40.21  & 19.94 & 30.66 & 2.11 ± 1.10 \\
+ DAMRO & 89.56 & 11.57 & 40.21 & 42.51  & 19.86 & 30.66 & 1.76 ± 1.05\\
+ PAI & 89.82 & 12.96 & 41.59 & 43.68  & 21.42 & 32.22 & 1.61 ± 1.04 \\
\rowcolor{aliceblue} Ours & 90.35 & 14.96 & 42.98 & 44.41  & 21.83 & 32.62 & 1.39 ± 0.91 \\
\bottomrule
\end{tabular} }
\end{table*}

\noindent
\textbf{Hallucination Mitigation on Open-Ended.} \\
We further evaluate several mitigation methods on the LLaVA-Med series, as presented in Table \ref{tab:type1_open_mitigation}. Certain techniques, such as PAI, which demonstrates strong performance on the close-ended benchmark, fail to generalize effectively to open-ended report generation. Some approaches reduce the hallucination rate (CHAIR) but do so at the expense of substantially lower recall, as observed with VCD and M3ID on LLaVA-Med. In contrast, other methods, such as DAMRO and OPERA on LLaVA-Med, improve both CHAIR and recall while also enhancing at least one report-specific metric.
Notably, our method exhibits the highest overall performance across
both architectural backbones. These results demonstrate that the proposed mitigation approach not only suppresses hallucinations more effectively than existing methods but also preserves factual completeness,
frequently enhancing it, while adhering to established radiology report evaluation criteria.

\subsection{Knowledge Deficiency Hallucination}
Hallucination can also occur when the model correctly interprets the image, such as recognizing key organs and visual features, but lacks the comprehensive medical knowledge required for accurate diagnosis or clinical decision.

\noindent
\textbf{Hallucination Mitigation on Closed-Ended. } \\
We further evaluate the effectiveness of various mitigation techniques. Unlike in visual-misinterpretation hallucination tasks, only our \ours\ and OPERA yield consistent performance improvements,
as shown in Figures \ref{fig:deficiency_a} and \ref{fig:deficiency_b}. Notably, PAI, which enhances attention over visual tokens and excels at reducing closed-ended visual hallucinations, fails to
address knowledge-level hallucinations (see Figure \ref{fig:deficiency_a}). This indicates that mitigating knowledge-based hallucinations requires strategies beyond purely visual enhancements.

\noindent
\textbf{Hallucination Mitigation on Open-Ended.} \\
According to Table \ref{tab:open_ended_KH_mitigation}, the application of mitigation techniques on the LLaVA-Med series yields only modest improvements for our approach. A limited number of methods, such as DoLa and PAI, reduce hallucinations across both backbones, whereas the majority of techniques result in increased hallucination scores.
Furthermore, the impacts of most mitigation strategies
vary between Med-LVLMs, revealing inconsistencies in their efficacy and underscoring the complexity of knowledge-deficiency hallucinations.

\subsection{Context Misalignment Hallucination}
In addition to the evaluations outlined in previous sections, clinical practice necessitates that medical image interpretation aligns with the patient’s comprehensive medical history. This includes
critical factors such as treatment plans, diagnostic records, family history, and other relevant clinical data. However, existing benchmarks for assessing hallucination in medical imaging predominantly
focus on isolated image analysis, neglecting the broader clinical context integral to real-world practice. To address this gap and better align with the practical demands of the medical field, we
evaluate the model’s susceptibility to hallucinations by contextualizing medical images within the patient’s holistic medical background.

\begin{figure*}[htp]
\centering
    \subfloat[Hallucination mitigation on LLaVA-Med]{%
        \includegraphics[width=0.48\linewidth]{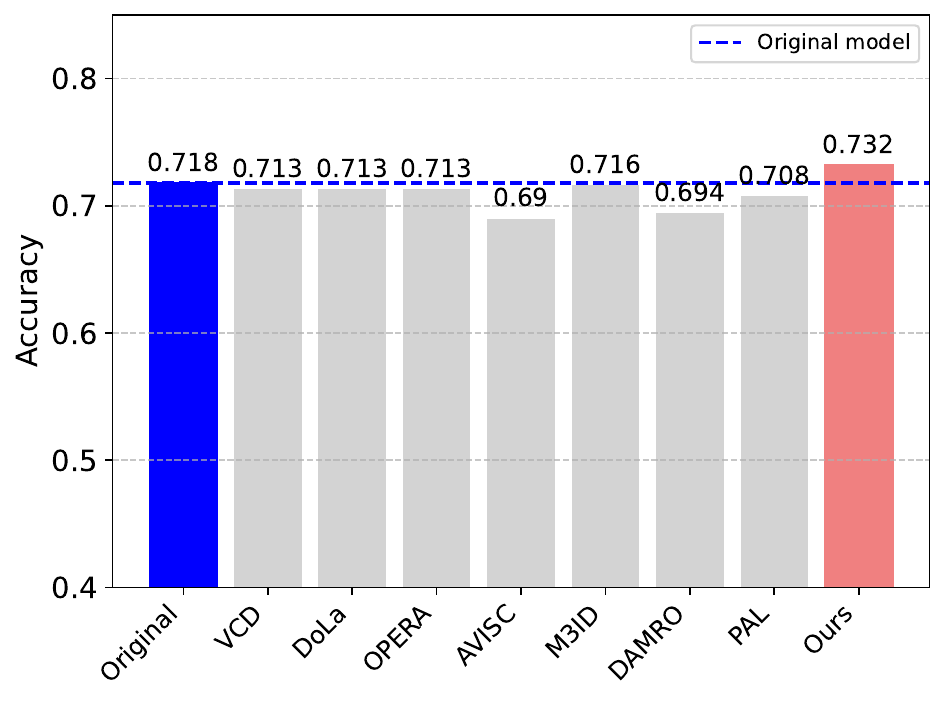}%
        \label{fig:misalignment_a}%
        }%
    \hfill%
    \subfloat[Hallucination mitigation on LLaVA-Med-1.5]{%
        \includegraphics[width=0.48\linewidth]{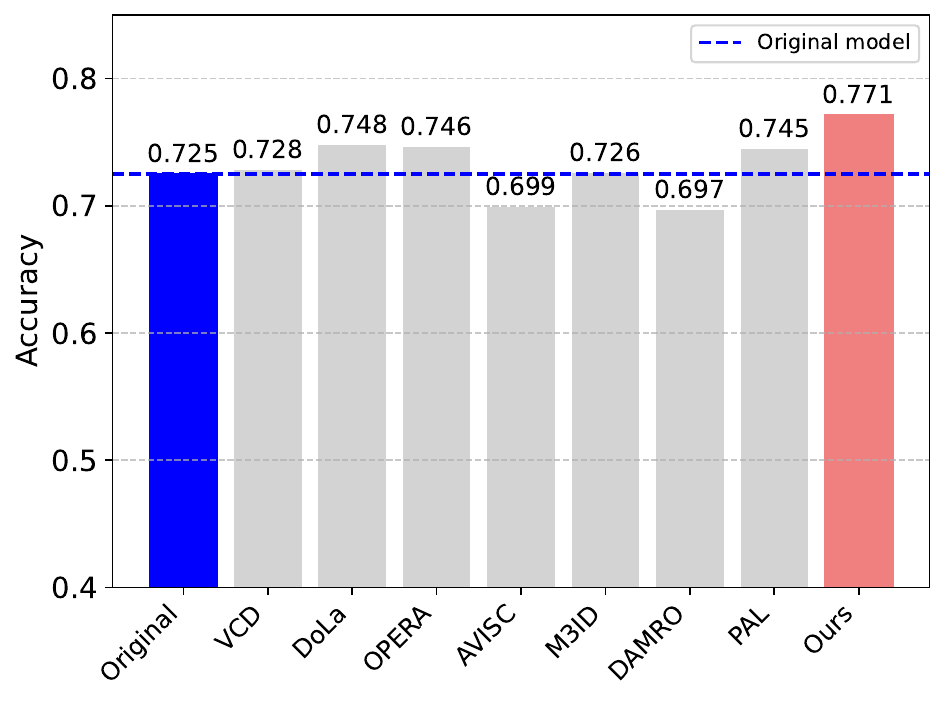}%
        \label{fig:misalignment_b}%
        }%
    \caption{Close-ended evaluation of context misalignment hallucination in (Med)-LVLMs and the effectiveness of hallucination
mitigation methods.}
\end{figure*}

\noindent
\textbf{Mitigation Evaluation Results.} \\
As illustrated in Figure \ref{fig:misalignment_a} and \ref{fig:misalignment_b}, existing hallucination mitigation methods achieve limited success. For instance, on the LLaVA-Med benchmark, only
our proposed method demonstrates meaningful performance improvement, underscoring the inadequacy of current approaches in addressing contextual hallucinations. While our method exhibits a marked
improvement in LLaVA-Med-1.5, alternative techniques such as OPERA, PAI, and DoLa yield only marginal gains. These improvements are notably less significant compared to their efficacy in mitigating visual hallucinations.

\begin{table}[htbp]
\centering
\begin{adjustbox}{width=\linewidth}
\begin{tabular}{l}
\hline
\textbf{Case 1: Hallucination and mitigation in MM-VisHal} \\ \hline
\begin{tabular}[c]{@{}l@{}}
\includegraphics[width=3cm]{images/MM.jpg}
\end{tabular}
\begin{tabular}[c]{@{}l@{}}
\textbf{Question:} ``Is there evidence of brain edema in the image?''\\
\textbf{Ground truth:} ``Yes.''\\
\textbf{LLaVA-Med:} {\color{red}``No, there is no evidence of edema}\\{\color{red} in the brain in the MRI.''}\\
\textbf{Ours:} \textcolor{ForestGreen}{``Yes, there is evidence of edema in brain tissue.''}
\end{tabular}\\
\hline
\textbf{Case 2: Hallucination and mitigation in CXR-VisHal} \\ \hline
\begin{tabular}[c]{@{}l@{}}
\includegraphics[width=3cm]{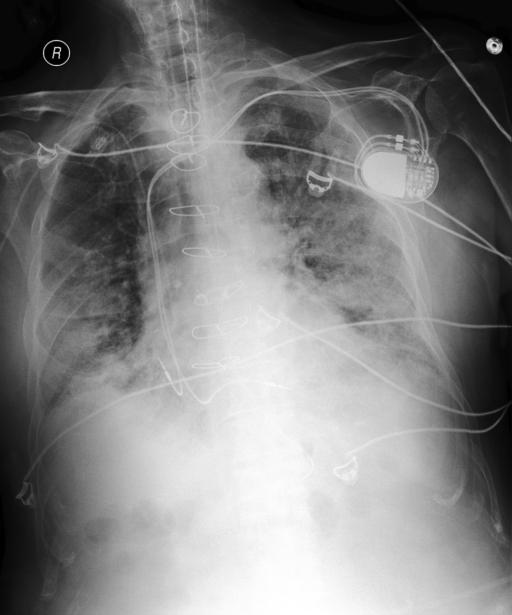}
\end{tabular}
\begin{tabular}[c]{@{}l@{}}
\textbf{Question:} ``Does the image show any signs of cardiomegaly?''\\
\textbf{Ground truth:} ``Yes.''\\
\textbf{LLaVA-Med:} {\color{red}``The chest X-ray does not show any signs of}\\{\color{red} cardiomegaly.''}\\
\textbf{Ours:} \textcolor{ForestGreen}{``Yes, the chest X-ray shows signs of}\\\textcolor{ForestGreen}{ cardiomegaly.''}
\end{tabular}\\
\hline
\end{tabular}
\end{adjustbox}
\caption{Examples of hallucination and mitigation cases.}
\label{tab:example}
\end{table}

\begin{table}[t]
\centering
\begin{adjustbox}{width=\linewidth}
\begin{tabular}{l}
\hline
\textbf{Case studies on hallucination and mitigation in report generation.} \\ \hline
\begin{tabular}[c]{@{}l@{}}
\includegraphics[width=3cm]{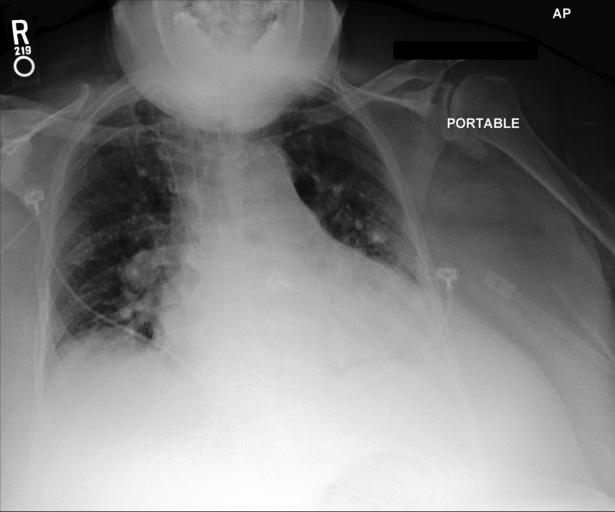}
\end{tabular}
\begin{tabular}[c]{@{}l@{}}
\textbf{Ground truth report:} ``The chest X-ray reveals an enlarged \\
cardiac silhouette, which is indicative of an increase in heart size. \\
There is also evidence of pulmonary edema, which is fluid \\
accumulation in the lungs. Additionally, a central catheter is\\
present, extending to the mid to lower superior vena cava (SVC).\\
The possibility of superimposed pneumonia is suggested \\
but not confirmed due to the absence of a lateral view.''\\[8pt]
\end{tabular}\\
\begin{tabular}[c]{@{}l@{}}
\textbf{LLaVA-Med}: ``Chest radiograph shows the \textcolor{ForestGreen}{cardiomediastinal silhouette}\\
is broadened, suggesting possible cardiomegaly or volume overload. Hilar and \\
mediastinal contours otherwise appear unremarkable. {\color{red}No pulmonary edema},\\
free intraperitoneal air, acute rib or clavicular fracture is identified. There \\
is no right internal jugular catheter visible,
and {\color{red}no pneumothorax} is observed.''\\[8pt]
\textbf{OPERA}: ``The chest X-ray shows relevant measurements including the volume\\
of the lungs, the size of the heart, and the position of the diaphragm. \\
Additionally, the image may show any abnormalities or signs of symptoms \\
related to the chest, such as \textcolor{ForestGreen}{enlarged cardiac silhouette}, fluid accumulation, \\
or \textcolor{ForestGreen}{superimposed pneumonia}.''\\[8pt]
\textbf{Ours}: ``The chest radiograph shows cardiomegaly, evidenced by an enlarged \\
\textcolor{ForestGreen}{cardiac silhouette}, along with radiographic signs of \textcolor{ForestGreen}{pulmonary edema}.\\
A central \textcolor{ForestGreen}{venous catheter} is visible, its tip projecting into the mid-to-lower\\
 portion of the superior vena cava.''\\[8pt]
\end{tabular}\\ \hline
\end{tabular}
\end{adjustbox}
\caption{Case studies highlighting visual misinterpretation hallucination and mitigation in medical report generation.}
\label{tab:report_example}
\end{table}

\subsection{Case Study for Hallucination and Mitigation}
\label{vfh_close_cases}
Table~\ref{tab:example} presents case studies from close-ended datasets on visual misinterpretation hallucinations. We provide the produced replies from our technique and baseline approach
in Cases 1 and 2 to visually show the efficacy of our strategy.
our method which enhances visual grounding through attention modification, successfully corrects the hallucinated responses. While it does not work for LLaVA-Med.
This study reinforces our findings that our mitigation method exhibit strength in hallucination mitigation, emphasizing the need for task-specific approach to improve Med-LVLM performance.

Also, Table~\ref{tab:report_example} shows case studies of open-ended report generation on visual misinterpretation hallucinations. In this example, our method effectively mitigates the hallucination
while even improving the recall of key findings. However, OPERA~\cite{huang2024opera}, while aiming to lower hallucination rates, significantly impact generation quality and recall, demonstrating the
challenges of balancing hallucination mitigation and report completeness.

\subsection{Ablation Studies.}
To thoroughly assess the efficacy of the elements of our suggested \ours\, we perform ablation experiments on the open-ended visual misinterpretation benchmark.
In particular, we assess the components' efficacy by eliminating or altering the particular settings, as indicated by the findings in Table~\ref{tab:ablation}.

\begin{table*}[htbp]
    \centering
    \caption{ Ablation experiments on the open-ended visual misinterpretation benchmark. Bold number indicates the best performance. VATS: Visual-Aware Token Selection;
    VPS: Visual Perception Score; MbS : Mask-based Sparsification; SAC : Sinking Attention Calibration. }
    \label{tab:ablation}%
             \resizebox{.9\textwidth}{!}{
      \begin{tabular}{cl|cccc|cccc}
      \toprule
      \multirow{2}[2]{*}{\textbf{ID.}} & \multicolumn{1}{l|}{\multirow{2}[2]{*}{\textbf{Settings}}} & \multicolumn{4}{c|}{\textbf{LLaVA-Med}} & \multicolumn{4}{c}{\textbf{LLaVA-Med-1.5}} \\
            &       & RadGraph $\uparrow$ & RaTEScore $\uparrow$ &  CHAIR $\downarrow$  & TPS $\uparrow$ & RadGraph $\uparrow$ & RaTEScore $\uparrow$ &  CHAIR $\downarrow$ & TPS $\uparrow$\\
      \midrule
      \multirow{2}[2]{*}{1} & w/o Whole VATS (i.e., Eq.~\ref{eq:select}) & 8.52  & 36.19 & 16.26  & 27.55 & 8.01 & 30.51 & 17.37 & 25.54 \\
            & w/o VPS in Eq.~\ref{eq:select} & 8.81  & 38.45  & 15.71 & 30.96 & 8.07 & 31.05 & 17.91 & 27.80 \\
      \midrule
      \multirow{2}[2]{*}{2} & w/o Whole SVCD (i.e., Eq.~\ref{eq:sparse-con}) & 7.88  & 38.38 & 17.93 & \textbf{35.83} & 7.79 & 30.93 & 17.79 & \textbf{30.68} \\
            & w/o MbS in Eq.~\ref{eq:sparse-con}& 8.48  & 38.12 & 15.36 & 30.30 & 8.06 & 31.26 & 18.12 & 27.47 \\
      \midrule
      3     & w/o SAC (i.e., Eq.~\ref{eq:penilty}) & 8.47  & 38.61 & 15.90 & 30.92 & 8.12 & 31.04 & 17.90 & 27.96\\
      \midrule
      \rowcolor{aliceblue} 4     & Our Full \ours & \textbf{8.97} & \textbf{41.33} & \textbf{15.02} & 30.87 & \textbf{8.29} & \textbf{40.76} & \textbf{17.05} & 27.73  \\
      \bottomrule
      \end{tabular}%
     }
  \end{table*}%

\noindent\textbf{Impact of the VATS.}~
Eliminating the VATS mechanism reduces both model accuracy and decoding speed (tokens per second, TPS), as shown by Group 1 in Table \ref{tab:ablation}. This decline indicates
that partially sparsifying the decoding sequence can alleviate linguistic bias in Med-LVLMs by reducing the influence of specific tokens during attention operation. Performance degrades still further when
the visual-perception score is removed. Collectively, these results highlight the effectiveness of our proposed VATS strategy.

\noindent\textbf{Impact of the Sparse-based VCD.}~
We exclude both the entire sparse-based visual contrastive decoding (SVCD) and the mask-based sparsification $S^{m}$ from Equation~\ref{eq:sparse-con} in order to assess the performance of our SVCD.
We see a notable drop in performance, as seen in Group 2 of Table~\ref{tab:ablation}, which further confirms the efficacy of our SVCD and mask-based sparsification approach.

\noindent\textbf{Impact of the Sinking Attention Calibration.}~
Additionally, we eliminated the SAC mechanism and see a further reduction in the method's ability to mitigate hallucinations.
This illustrates the usefulness of the suggested attention calibration technique as well as the connection between hallucinations and sinking attention.

\begin{figure}
    \centering
    \begin{subfigure}{0.49\linewidth}
      \includegraphics[width=1\linewidth]{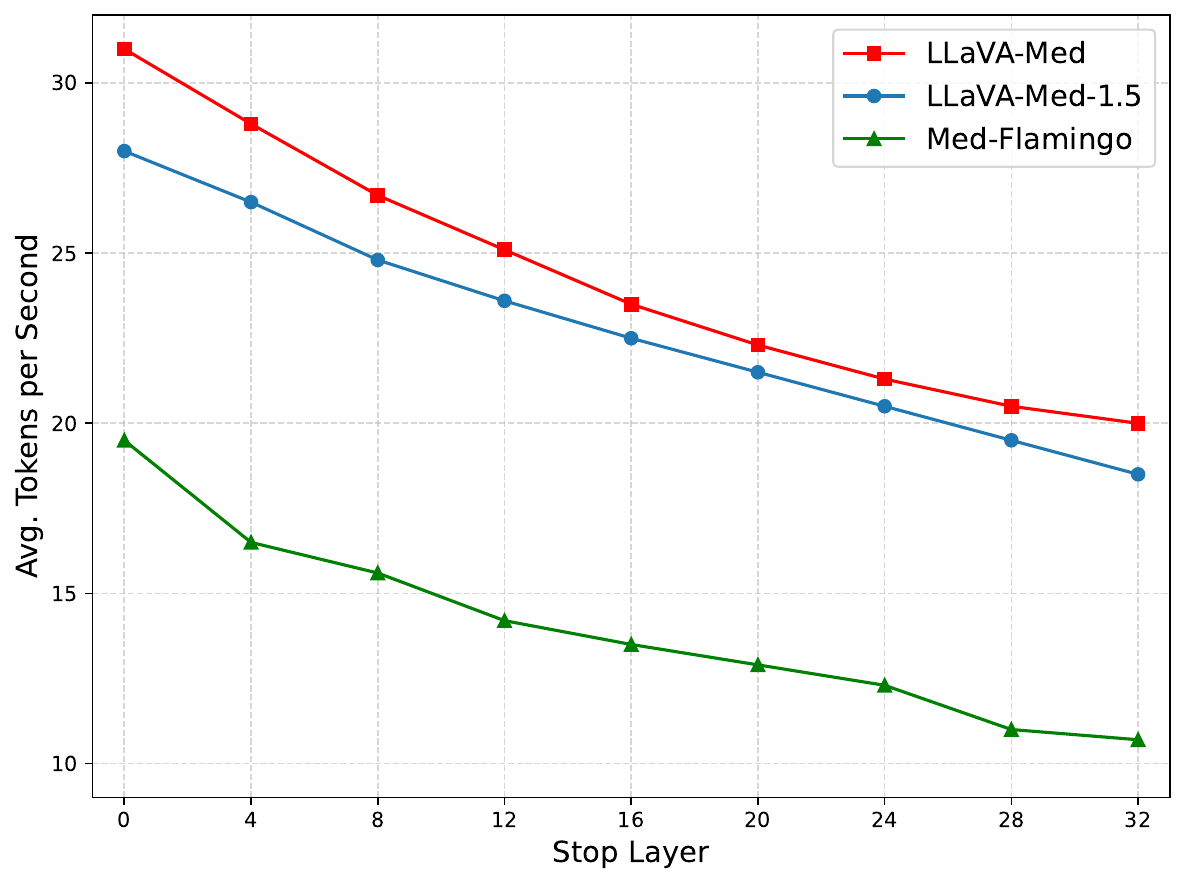}
      \caption{TPS during decoding.}
      \label{fig:exp-1}
    \end{subfigure}
    \hfill
    \begin{subfigure}{0.49\linewidth}
      \includegraphics[width=1\linewidth]{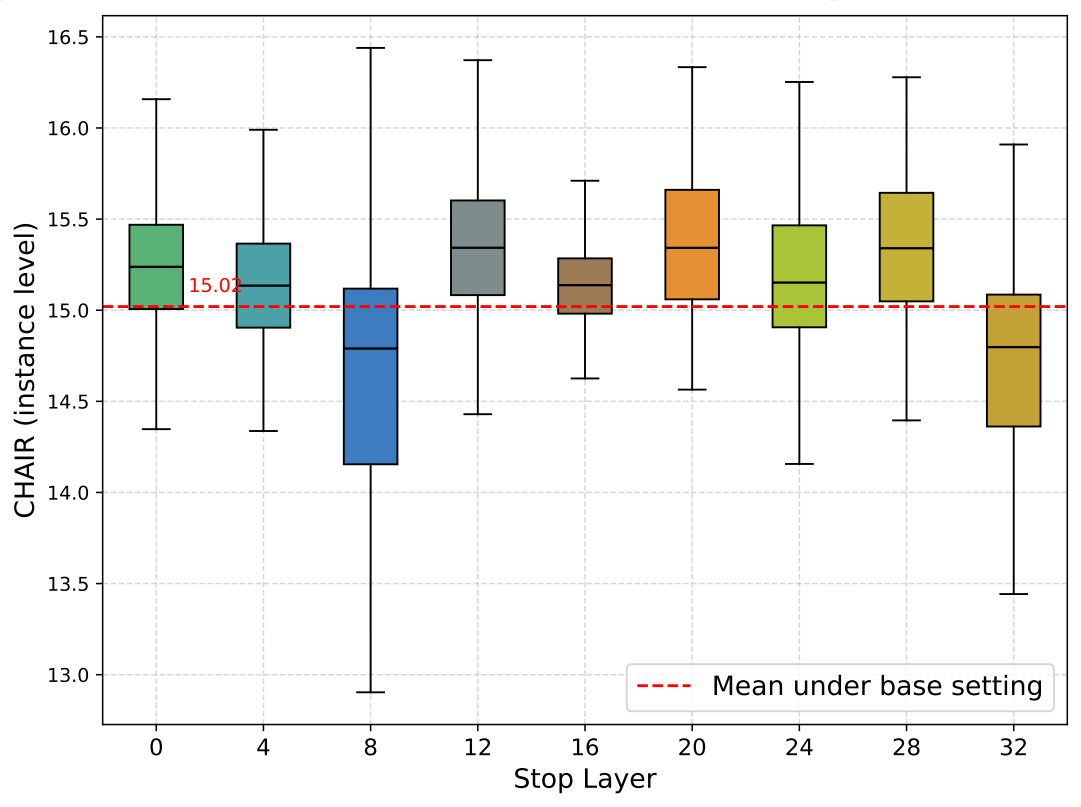}
      \caption{CHAIR evaluation results.}
      \label{fig:exp-2}
    \end{subfigure}
    \caption{Analysis of the effectiveness and performance of various logit sources: (a) how using various early stopping layers affects token per second (TPS); (b) how using various early stopping layers affects LLaVA-Med-1.5 performance.}
        \label{fig:exp}
\vspace{-0.3em}
  \end{figure}

\noindent\textbf{Decoding Efficiency Analysis.}~
As seen in Figure~\ref{fig:exp}, we compute the contrastive logits from features at various levels of the Med-LVLM decoder to calibrate the distribution and illustrate decoding speed and
performance in order to further evaluate the impact of applying embedding features to compute the proposed SVCD.
We find that our approach already provides good hallucination preventing performance while achieving optimal decoding speed when just embedded features are used (i.e., stop layer is 0).
Our \ours\ successfully prevents the time-consuming secondary decoding procedure in this manner, striking a compromise between efficiency and performance.

\noindent\textbf{Hyperparameter Sensitivity.}
We examine the robustness of \ours\ with respect to three key hyperparameters---$\alpha$, $\beta$, and $\lambda$---which regulate different components of the visual contrastive decoding framework on LLaVA-Med 7B. The results, summarized in Tables~\ref{tab:ablation2}, \ref{tab:ablation3}, \ref{tab:ablation1}, reveal consistent and interpretable trends across all evaluation metrics (RadGraph, RaTEScore, and CHAIR).

\noindent\textbf{Effect of $\alpha$.}
As shown in Table~\ref{tab:ablation2}, performance improves steadily as $\alpha$ increases from 0.1 to 0.3, peaking at $\alpha=0.3$ with the highest RadGraph (8.97) and RaTEScore (41.33) and the lowest CHAIR (15.02). Beyond this point, all metrics gradually degrade as $\alpha$ continues to increase, suggesting that excessive emphasis on the contrastive regularization term diminishes its discriminative benefit and introduces noise into the visual-text alignment process. The results indicate that a moderate contrastive weight ($\alpha \approx 0.3$) achieves the optimal balance between enforcing meaningful visual-text distinctions and maintaining stable generation quality.

\noindent\textbf{Effect of $\beta$.}
Similarly, Table~\ref{tab:ablation3} demonstrates a clear non-linear sensitivity to $\beta$, which controls the strength of the sparsification regularizer. The model achieves its best results at $\beta=0.1$, where RadGraph (8.97) and RaTEScore (41.33) reach their maximums and CHAIR (15.02) reaches its minimum. Smaller $\beta$ values yield under-regularized outputs with residual visual noise, whereas larger $\beta$ values overly penalize the regularizer, suppressing informative visual cues. This again supports that moderate sparsification encourages the model to focus on salient, contextually aligned visual tokens while avoiding over-filtering.

\noindent\textbf{Effect of $\lambda$.}
A similar trend is observed for $\lambda$, as shown in Table~\ref{tab:ablation1}, which modulates the contrastive loss scaling. Performance increases as $\lambda$ rises from 0 to 0.1, reaching the same optimal point at $\lambda=0.1$, before gradually declining for larger values. The consistent behavior across all three metrics confirms that an appropriate contrastive weighting fosters stable alignment and effective hallucination mitigation, while excessive scaling introduces imbalance between visual and textual modalities.

\begin{table}[ht!]
\centering
\resizebox{0.45\textwidth}{!}{
\begin{tabular}{lccc}
\midrule
\textbf{Method}  & \textbf{VQA-RAD} & \textbf{SLAKE} & \textbf{PathVQA} \\
\midrule
\multicolumn{4}{c}{\textbf{Clean}} \\
\midrule
PubMedCLIP        & 80.00 & 82.50 & 87.34 \\
\quad + Ours      & \textbf{82.10} & \textbf{84.35} & \textbf{88.90} \\
\midrule
BiomedCLIP        & 79.80 & 89.70 & 86.89 \\
\quad + Ours      & \textbf{81.63} & \textbf{91.17} & \textbf{88.42} \\
\midrule
BiomedGPT         & 57.84 & 73.30 & 84.20 \\
\quad + Ours      & \textbf{60.53} & \textbf{75.09} & \textbf{85.61} \\
\midrule
CLIP-ViT          & 81.15 & 82.10 & 87.13 \\
\quad + Ours      & \textbf{83.05} & \textbf{83.90} & \textbf{88.75} \\
\midrule
LLaVA-Med         & 84.19 & 85.34 & 91.21 \\
\quad + Ours      & \textbf{85.90} & \textbf{87.38} & \textbf{92.59} \\
\midrule
\multicolumn{4}{c}{\textbf{Adversarial}} \\
\midrule
PubMedCLIP        & 72.50 & 75.45 & 80.12 \\
\quad + Ours      & \textbf{75.08} & \textbf{77.64} & \textbf{82.10} \\
\midrule
BiomedCLIP        & 71.60 & 81.20 & 79.45 \\
\quad + Ours      & \textbf{74.00} & \textbf{83.00} & \textbf{81.30} \\
\midrule
BiomedGPT         & 50.36 & 66.81 & 77.14 \\
\quad + Ours      & \textbf{53.22} & \textbf{68.94} & \textbf{78.64} \\
\midrule
CLIP-ViT          & 73.20 & 74.56 & 79.61 \\
\quad + Ours      & \textbf{75.40} & \textbf{76.38} & \textbf{81.16} \\
\midrule
LLaVA-Med         & 77.09 & 78.10 & 84.30 \\
\quad + Ours      & \textbf{79.26} & \textbf{80.25} & \textbf{85.94} \\
\midrule
\end{tabular}
}
\caption{Performance on Med-VQA tasks under clean and adversarial settings. “+Ours” indicates our enhancement over the baseline models.}
\label{tab:main_table}
\end{table}

\begin{table*}[htbp]
    \centering
    \footnotesize
    \caption{Accuracy (\%) of different LVLMs and Med-LVLMs on medical VQA task. }
    \resizebox{\linewidth}{!}{
    \begin{tabular}{l|cc|c|cc}
    \toprule
        \multirow{2}{*}{Models} & \multicolumn{2}{c|}{\textbf{Radiology}} &\textbf{Ophthalmology} & \multicolumn{2}{c}{\textbf{Pathology}} \\
        & IU-Xray & MIMIC-CXR & Harvard-FairVLMed & Quilt-1M & PMC-OA (Pathology) \\ \midrule
        LLaVA-Med-1.5 & 75.47 & 75.79 & 63.03& 62.80& 59.28  \\
        +\ours\ &
        \textbf{90.56}& \textbf{82.95} &
        \textbf{88.73} &
        \textbf{73.52} & \textbf{65.31} \\ \midrule
        Med-Flamingo & 26.74 & 61.27 & 42.06& 27.11& 32.62 \\
        +\ours\ & \textbf{59.15} & \textbf{63.74} & \textbf{49.72} & \textbf{36.76} & \textbf{39.54} \\
        \midrule
        InternVL-2 & 54.06 & 59.47 & 44.38 & 37.82 & 34.40 \\
        +\ours\ &    \textbf{63.25} & \textbf{60.61} & \textbf{61.50} & \textbf{53.56} & \textbf{49.70} \\ \midrule
        Qwen-VL-Chat & 59.43 & 60.43 & 38.06 & 28.74 & 29.53 \\
        +\ours\ &     \textbf{73.34} & \textbf{66.06} & \textbf{39.92} & \textbf{31.81} & \textbf{29.77} \\
    \bottomrule
    \end{tabular}
    }
    \label{tab:app_vqa}
\end{table*}

\begin{table}[t]
\centering
\resizebox{.9\linewidth}{!}{%
\begin{tabular}{@{}c|ccc@{}}
\toprule
\textbf{$\alpha$} & RadGraph $\uparrow$ & RaTEScore $\uparrow$ & CHAIR $\downarrow$ \\ \midrule
0.1 & 8.42 & 39.85 & 15.47 \\
0.2 & 8.73 & 40.92 & 15.18 \\
0.3 & \textbf{8.97} & \textbf{41.33} & \textbf{15.02} \\
0.4 & 8.86 & 41.10 & 15.10 \\
0.5 & 8.65 & 40.60 & 15.25 \\
0.6 & 8.40 & 39.95 & 15.40 \\
0.7 & 8.15 & 39.10 & 15.65 \\
0.8 & 7.95 & 38.55 & 15.80 \\
0.9 & 7.82 & 38.10 & 15.92 \\
1.0 & 7.70 & 37.80 & 16.00 \\
\bottomrule
\end{tabular}
}
\caption{An ablation study of $\alpha$ on the open-ended visual misinterpretation benchmark.}
\label{tab:ablation2}
\end{table}

\begin{table}[t]
\centering
\resizebox{.9\linewidth}{!}{%
\begin{tabular}{@{}c|ccc@{}}
\toprule
\textbf{$\beta$} & RadGraph $\uparrow$ & RaTEScore $\uparrow$ & CHAIR $\downarrow$ \\ \midrule
0     & 8.38 & 39.70 & 15.58 \\
0.001 & 8.62 & 40.55 & 15.30 \\
0.01  & 8.80 & 40.95 & 15.15 \\
0.1   & \textbf{8.97} & \textbf{41.33} & \textbf{15.02} \\
0.2   & 8.90 & 41.18 & 15.08 \\
0.3   & 8.84 & 41.00 & 15.14 \\
0.4   & 8.75 & 40.85 & 15.20 \\
0.5   & 8.63 & 40.60 & 15.28 \\
0.6   & 8.52 & 40.30 & 15.35 \\
0.7   & 8.43 & 40.05 & 15.42 \\
0.8   & 8.35 & 39.85 & 15.48 \\
0.9   & 8.30 & 39.75 & 15.52 \\
1.0   & 8.25 & 39.65 & 15.57 \\
\bottomrule
\end{tabular}
}
\caption{An ablation study of $\beta$ on the open-ended visual misinterpretation benchmark.}
\label{tab:ablation3}
\end{table}

\begin{table}[t]
\centering
\resizebox{.9\linewidth}{!}{%
\begin{tabular}{@{}c|ccc@{}}
\toprule
\textbf{$\lambda$} & RadGraph $\uparrow$ & RaTEScore $\uparrow$ & CHAIR $\downarrow$ \\ \midrule
0     & 8.40 & 39.85 & 15.50 \\
0.001 & 8.68 & 40.75 & 15.25 \\
0.01  & 8.85 & 41.05 & 15.12 \\
0.1   & \textbf{8.97} & \textbf{41.33} & \textbf{15.02} \\
0.2   & 8.92 & 41.20 & 15.07 \\
0.3   & 8.86 & 41.05 & 15.12 \\
0.4   & 8.78 & 40.90 & 15.18 \\
0.5   & 8.68 & 40.65 & 15.25 \\
0.6   & 8.55 & 40.35 & 15.32 \\
0.7   & 8.45 & 40.10 & 15.38 \\
0.8   & 8.37 & 39.90 & 15.45 \\
0.9   & 8.32 & 39.80 & 15.50 \\
1.0   & 8.28 & 39.70 & 15.55 \\
\bottomrule
\end{tabular}
}
\caption{An ablation study of $\lambda$ on the open-ended visual misinterpretation benchmark.}
\label{tab:ablation1}
\end{table}

\noindent\textbf{Evaluation on Large Vision-Language Models.}
Table~\ref{tab:main_table} reports results on three Med-VQA datasets (VQA-RAD, SLAKE, and PathVQA) under both clean and adversarial evaluation conditions. Across all base models, including PubMedCLIP, BiomedCLIP, BiomedGPT, CLIP-ViT, and LLaVA-Med, integrating \ours\ yields consistent accuracy improvements.
Under clean settings, \ours\ achieves gains of up to $+2.0\%$ absolute improvement on average, demonstrating more accurate and visually grounded reasoning.
Importantly, under adversarial perturbations that intentionally distort visual cues, \ours\ remains robust, outperforming baselines by a similar or even larger margin.
This robustness indicates that the proposed contrastive regularization effectively strengthens visual-text alignment and mitigates susceptibility to misleading or corrupted image signals, a critical property for trustworthy medical reasoning systems.

\noindent\textbf{Cross-Domain Generalization.}
Table~\ref{tab:app_vqa} further validates the scalability of \ours\ across different medical imaging modalities and LVLM architectures, including LLaVA-Med-1.5, Med-Flamingo, InternVL-2, and Qwen-VL-Chat.
The performance consistently improves across radiology (IU-Xray, MIMIC-CXR), ophthalmology (Harvard-FairVLMed), and pathology (Quilt-1M, PMC-OA) datasets.
For example, \ours\ elevates LLaVA-Med-1.5 accuracy from 75.47\% to 90.56\% on IU-Xray and from 63.03\% to 88.73\% on Harvard-FairVLMed, reflecting substantial gains in visual grounding and reasoning precision.
Similar trends hold for other foundation models, even those with limited domain supervision such as Med-Flamingo and InternVL-2, confirming that \ours\ enhances both medical-specific and general-purpose LVLMs without retraining.

\begin{table}[t]
\centering
\caption{Efficiency comparison on LLaVA-Med using the CXR-VisHal dataset with an NVIDIA A100.}
\label{table_e}
\resizebox{0.9\columnwidth}{!}{
\begin{tabular}{lccc}
    \hline
    Methods  & \textbf{Time(s)} $\downarrow$ & \textbf{Memory(MB)}$\downarrow$ & \textbf{Accuracy(\%)}$\uparrow$ \\ \hline
    Normal  & 494  & 15673  & 69.8 \\
    VCD      & 904  & 16753  & 69.7     \\
    DoLa      & 974  & 16843  & 72.3     \\
    PAI      & 1974  & 19286  & 72.6     \\
    OPERA    & 2643 & 21943  & 73.2       \\
    \textbf{\ours} & 936  & 16987  & 75.7     \\
    \hline
    \end{tabular}}
\end{table}

\noindent\textbf{Computation Efficiency.}
One major challenge for hallucination-mitigation decoding methods lies in their computational overhead. To quantify this, we evaluate the total inference time (in seconds) and peak GPU memory consumption (in MB) of LLaVA-Med-7B on the CXR-VisHal dataset, as reported in Table~\ref{table_e}. The Contrastive Decoding method (VCD) requires generating distorted raw inputs, which effectively doubles the inference complexity. OPERA relies on beam-search decoding and maintains multiple candidate beams, while its retrospection–reallocation strategy introduces an additional rollback mechanism, further compounding computational costs. In contrast, \ours\ mitigates vision–text association hallucinations by pruning a large proportion of attention-critical tokens in the early layers, thereby achieving a favorable trade-off between accuracy and efficiency, with an inference time of 936 seconds.

\section{Conclusion}
This paper introduces \ours, a visual contrastive decoding system designed to mitigate visual hallucination issues in Med-LVLMs. We first conduct an in-depth analysis of how visual-agnostic token
sparsification exacerbates hallucinations in model outputs. Building on this analysis, we present a visual-aware token selection procedure to refine the decoding process by prioritizing tokens with
stronger relevance to the input visual data. Furthermore, we introduce a sparsity-based visual contrastive decoding framework that recalibrates logits directly during inference, eliminating the
need for secondary decoding while optimizing the model’s attention mechanism. Experiments across diverse medical imaging domains and benchmarks demonstrate that \ours\ significantly enhances factual
accuracy and reduces hallucination rates. These results underscore the potential of our approach to improve the reliability of Med-LVLMs in clinical applications, ensuring their utility as trustworthy
tools for healthcare decision-making. For future work, we plan to explore complementary directions such as fine-tuning and knowledge-augmented methods to further enhance LVLM performance. In addition, developing dedicated hallucination evaluation datasets for models like MedCLIP and MedSAM will be essential to broaden benchmarking capabilities and foster more robust comparisons across architectures.

\bibliographystyle{elsarticle-num.bst}
\bibliography{refs}

%


\end{document}